\documentclass[10pt,journal,compsoc]{IEEEtran}
\usepackage[nocompress]{cite}
\usepackage[pdftex]{graphicx}
\usepackage{balance}
\graphicspath{{figures/}}
\DeclareGraphicsExtensions{.pdf,.jpeg,.png}
\usepackage{times}
\usepackage{epsfig}
\usepackage{amsmath}
\usepackage{amssymb}
\usepackage{overpic}
\usepackage{subfigure}
\usepackage{multirow}
\usepackage{color}
\usepackage{diagbox}
\usepackage{mathtools}
\usepackage{bbding}
\usepackage{booktabs}

\newtheorem{axiom}{Axiom}

\newcommand{\wh}[1]{\textcolor{black}{#1}}

\usepackage[colorlinks]{hyperref}

\usepackage{color,soul}
\usepackage[table]{xcolor}
\usepackage[switch]{lineno}

\renewcommand{\raggedright}{\leftskip=0pt \rightskip=0pt plus 0cm}

\begin{document}

\title{Coding for Intelligence from the Perspective \\ of Category}
\author{Wenhan Yang, Zixuan Hu, Lilang Lin, 
        Jiaying Liu, Ling-Yu Duan
        \IEEEcompsocitemizethanks{
        \IEEEcompsocthanksitem Wenhan Yang, Zixuan Hu, Lilang Lin, Jiaying Liu, and Ling-Yu Duan are with Peking University, Beijing 100871, China. E-mail: 
        \{yangwenhan, hzxuan, linlilang, liujiaying, lingyu\}@pku.edu.cn.
		}
}
\markboth{}
{Shell \MakeLowercase{\textit{et al.}}: Bare Demo of IEEEtran.cls for Computer Society Journals}

\IEEEtitleabstractindextext{

\begin{abstract}
\raggedright 
Coding, which targets compressing and reconstructing data, and intelligence, often regarded at an abstract computational level as being centered around model learning and prediction, interweave recently to give birth to a series of significant progress.
The recent trends demonstrate the potential homogeneity of these two fields, especially when deep-learning models aid these two categories for better probability modeling.
For better understanding and describing from a unified perspective, inspired by the basic generally recognized principles in cognitive psychology,
we formulate a novel problem of \textit{Coding for Intelligence} (CfI) from the category theory view.
Based on the three axioms: \textit{existence of ideal coding}, \textit{existence of practical coding}, and \textit{compactness promoting generalization}, we derive a general framework to understand existing methodologies, namely that, coding captures the intrinsic relationships of objects as much as possible, while ignoring information irrelevant to downstream tasks.
This framework helps identify the challenges and essential elements in solving the specific derived Minimal Description Length (MDL) optimization problem from a broader range, providing opportunities to build a more intelligent system for handling multiple tasks/applications with coding ideas/tools.
Centering on those elements, we systematically review recent processes of towards optimizing the MDL problem in more comprehensive ways from data, model, and task perspectives, and reveal their impacts on the potential CfI technical routes.
After that, we also present new technique paths to fulfill CfI and provide potential solutions with preliminary experimental evidence.
Via organizing bitstreams of multiple tasks and multi-modality data in a scalable way or parameterized as new latent variables,
the improved performance for analytics tasks is observed and it is seen that such joint optimization is well aligned with the modeling of large foundation models.
Last, further directions and remaining issues are discussed as well.
The discussion shows our theory can reveal many phenomena and insights about large foundation models, which mutually corroborate with recent practices in feature learning.
\end{abstract}
}
     
\maketitle



\section{Introduction}
The puzzle of what intelligence is has always been a fundamental scientific question in the long exploration history of humans themselves.
Despite continuous investigations from various disciplines, it is still an open issue with no visible solution or clear path to solve today.
Cognitive psychology~\cite{sternberg2011cognitive} makes efforts to uncover the fundamental physiological mechanisms of intelligence, \textit{e.g.} abstraction~\cite{tyler2010psy}, analogy~\cite{dedre1983analogy}, and generalization~\cite{banich2010generalization} from the perspective of understanding the key processes in human cognition, such as memory~\cite{brady2008memory}, knowledge~\cite{birbaumer1999spelling}, problem-solving~\cite{reed2000enc}, and decision~\cite{johnson2000reason}.
This contributes to a deeper understanding of the essence of intelligence in both theoretical foundations and empirical evidence.
Artificial intelligence (AI)~\cite{russell2016artificial,dayan2005neuroscience}, from a computational standpoint, simulates and carries out tasks to simulate human intelligence, to imitate human cognitive processes through algorithms and computer programs.

The study of neuroscience~\cite{squire2012neuroscience} focuses on understanding the structure and function of the nervous system, especially the brain, to reveal the biological basis of intelligent behavior.
The way humans process information serves as guidance to inspire the computational implementation of artificial intelligence.
As demonstrated in~\cite{roozendaal2002neurobiology,roozendaal2003neuro,metcalfe1986feelingok}, coding and memory mechanisms play a critical role in creating intelligence.
By compressing and reorganizing information, short-term memory can transform into long-term memory easily retrievable and usable.
This biological evidence well aligns with the recent results in computer vision, which demonstrates that data compression and task-driven data organization contribute to realizing intelligence.
Namely, coding is one of the most critical factors that leads to intelligence computationally.

From a historical standpoint, coding is in fact always closely related to intelligent technology.
On one hand, coding enables data interchange and makes computations of models more efficient.
Additionally, entropy, which plays the role of information bottleneck to preserve intrinsic semantics while removing redundancy, is inspiring the design of machine learning models.
For example, a significant amount of research on deep learning theories and methods adopts explicit or implicit information bottlenecks (\textit{e.g.} dropout~\cite{srivastava2014drouput}, variational auto-encoder~\cite{doersch2016vae}), fully demonstrating the important role that insights from coding theory play in improving model learning's capacity.
On the other hand, intelligent technology, which drives the models to provide stronger capacities, also promotes the progress of coding.
The emergence of predictive coding and hybrid coding~\cite{ma2015avs2,bross2021vvc} both validate the utilization of fine-grained models and more diverse patterns to improve coding efficiency.

The rapid development of deep learning has triggered a new revolution of promoting and fusing these two fields with increasingly powerful models and more flexible training methods.
In computer vision, significant breakthroughs are achieved in a series of tasks such as classification~\cite{simonyan2015very,he2016resnet}, detection~\cite{girshick2015fastrcnn,redmon2016yolo}, segmentation~\cite{long2015fcn}, retrieval~\cite{noh2017retrieval}, \textit{etc}.
The entropy bottleneck mechanism becomes a central concept for measuring and constraining features from a general sense.
In video compression, the coding performance continuously improves by using more complicated data-dependent modes or patterns and end-to-end optimization.

In recent years, generative models/AI and large foundation models~\cite{touvron2023llama,kirillov2023segany,evaclip} witness numerous successful applications, capable of accurately characterizing joint probability distributions. 
Thus, with proper prompt cues~\cite{huang2023densityprompt,liu2023explicit}, the flexible sampling is easily obtained from the conditional probability, for excellent quality that satisfies either semantics or human visual perception.
This capacity comes from the essential compression inherent nature~\cite{deletang2023language} in large foundation models.
When all data distributions are compactly represented, the semantic representation and signal reconstruction can be correlated, which helps offer their effective joint optimization for both reconstruction and analytics.

In this context, the field of coding has also strived to explore a broader range of coding issues.
As early as 2020, the paradigm of Video Coding for Machines~\cite{duan2020vcm,hu2020vcm,yang2024compresstaskonomy} has been proposed for flexible joint optimization of multiple tasks (integrating both low-level signal reconstruction and high-level semantic analytics) in a scalable manner with relatively small communication and computation costs.
When these efforts intersect with large foundation models and collaborative infrastructure, we believe that the integration of diverse semantic representation with signal compression/reconstruction into a compact representation will be obtained, \textit{i.e.} compressing everything (different objects, modalities, tasks, \textit{etc.}) for modeling anything.
Therefore, compared with ever before, we are capable of bridging the independent development paths of coding in specific signal domains and prediction/decision models in the field of AI.

Now, standing at the critical point of the preliminary integration of coding and intelligent technologies, their deep joint optimization still lacks in both theory and practice.
%
%
%
In this paper, we start by revisiting the classic principles of cognitive psychology, which helps us establish a few axioms from the well-accepted basis of category theory and Kolmogorov complexity.
Based on these axioms, we further build a novel coding paradigm that reorganizes the efforts of all previous coding routes in a unified manner.
With the help of category theory, we obtain the formulation of how to build an intelligent system from the coding view in a general way, unifying the statement about several mainstream computational problems related to AI.
Several instances are given to describe the specific intelligent coding problem, such as reconstruction-driven methods (including image compression, and super-resolution) 
and understanding-driven methods (including feature compression, and fully supervised learning).
Then, by specifying and simplifying the formulation, we further reach a linear combination form, from which we summarize previous related technical routes and identify several important technical trends that help promote intelligence with the help of coding ideas/tools.
Our contributions are summarized as follows,
\begin{itemize}
    \item We formulate the novel problem of Coding for Intelligence from the category view in the general sense, which helps identify the challenges and essential elements in solving the related optimization problem from a broader range, providing opportunities to build a more intelligent system for handling multiple tasks/applications.
    \item We review the state-of-the-art approaches in both interactions between coding and intelligence techniques as well as the critical technical components related to CFI. 
    We also study the impact of recent technical evolution in terms of model, data, and task on the potential CFI technical routes.
    \item We present new technique paths that have the potential to lead to CFI architectures and provide potential solutions. Preliminary experimental results have shown the advantages of related components in the effectiveness of using compact representation for coding for human/machine vision, demonstrating their great potential for CFI.
\end{itemize}

The rest of the article is organized as follows. 
Section~\ref{sec:for} provides the mathematical definition and formulation of CfI from the category theory perspective which offers a more general view to look into both coding and AI techniques.
Section~\ref{sec:ins} briefly reviews several instances under the CfI scope.
Section~\ref{sec:sum} illustrates the emerging techniques centering on CfI from the perspective of data, model, and task.
Section~\ref{sec:trend} illustrates the emerging related solutions with preliminary experimental results, which provides useful evidence for CfI.
In Section~\ref{sec:diss}, several issues, and future directions are discussed. 
In Section~\ref{sec:con}, the concluding remarks are provided.

\section{Formulation}
\label{sec:for}
This section begins by exploring cognitive psychology to derive the fundamental elements of intelligence.
Then, based on these elements, we employ category theory to obtain the most abstract description, 
leading to the axiomatization of coding for intelligence, and finally presenting the relevant definitions.

\subsection{Psychology Basis}
``What intelligence is'' has always been one of the fundamental questions that cognitive psychology seeks to investigate.
Despite nearly a century of research, answers to this question have not converged.
In a recent work, Legg and Hutter cited about 70 definitions~\cite{legg2007ui}.
In our work, we do not attempt to provide a specific definition,
but rather systematically identify commonalities among these different perspectives,
thereby formalizing the fundamental elements of intelligence.

Among all intelligence models, there are generally recognized ones, such as
Carroll's three-stratum model~\cite{bickley1995threestratum},
theory of multiple intelligence~\cite{davis2011mi} proposed by Gardner, 
and triarchic theory of human intelligence~\cite{sternberg1985biq} proposed by Sternberg, \textit{etc}.
Among these classic models, three functions are generally expected and required:
\begin{itemize}
    \item \textbf{Analogy}: comparison between two things or two sets considering their common/shared elements~\cite{hofstadter2001analogy};
    \item \textbf{Abstraction}: derivation of general rules and concepts from specific examples~\cite{Wiki2024abs};
    \item \textbf{Generalization}: formulation of common properties of specific instances as general concepts~\cite{gluck2007learning}.
\end{itemize}

To enable these targets, information has to be recorded and processed as two forms~\cite{sternberg2011cognitive}:
1) \textit{Memory}:
Via efficient processing and \textbf{coding for storage}, information can be saved in the \wh{human brain} for fast extraction and retrieval;
By \textbf{coding to mine the correlation} between new information and past memory, short-term memory can be converted to the long one.
2) \textit{Knowledge}: In the process of information/memory turning to knowledge, an intrinsic point is to create/describe the \textbf{relationship}. For example, descriptive knowledge relies on features/prototypes~\cite{clark1977psychology,medin1981}, and builds hierarchical networks~\cite{collins1969}, or compares semantic features~\cite{smith1974}.
The integration of descriptive knowledge and procedural knowledge can be simulated by parallel distributed processing modules~\cite{Rogers2004}, \textit{i.e.} \textbf{connectionist} models.

From this, it can be seen that coding is an important aspect of information processing, while relationships constitute a critical core of information/knowledge.
The process of information being transformed into memory and knowledge, facilitates and reinforces the functionalities mentioned above. For example, the description of relationships promotes the generation of analogies, while connectionism stimulates the processes of generalization and abstraction.
Therefore, in the next subsection, we will start with relationships and coding as the most elementary basis, 
introducing category theory as a framework with the broad descriptive power to capture relationships and unfold our modeling and definition derivation.

\subsection{A Category Theory Perspective}
\vspace{1mm}

\noindent \textbf{Basics of Category Theory}. Following~\cite{yuan2023categorical}, we define several basic concepts of category theory:

\renewcommand{\baselinestretch}{1.06}

\begin{itemize}
    \item A category $\mathcal{W}$ is a quadruple $(\text{Ob}(\mathcal{W}), \text{Hom}_{\mathcal{W}}, \circ, \text{id})$, consisting of: a set of objects $\text{Ob}(\mathcal{W})$, a set of morphisms $\text{Hom}_{\mathcal{W}}$, a composition operation $\circ$, and a set of identity morphisms $\text{id}$;
    \vspace{1.5mm}
    \item For each pair $X, Y \in \text{Ob}(\mathcal{W})$, a set $\text{Hom}_{\mathcal{W}}(X, Y) \subset \text{Hom}_{\mathcal{W}}$ are called the $\mathcal{W}$-morphism from $X$ to $Y$. \vspace{1.5mm}
    \item Given $f \in \text{Hom}_{\mathcal{W}}(X, Y), g \in \text{Hom}_{\mathcal{W}}(Y, Z)$, we define the composition operation as $g\circ f \in \text{Hom}_{\mathcal{W}}(X, Z)$. The composition $\circ$ is associative:  $(  h \circ g ) \circ f = h \circ ( g \circ f)$; \vspace{1.5mm}
    \item For every $X \in \text{Ob}(\mathcal{W})$, there exists an identity morphism with respect to composition $\text{id}_{X}\in \text{Hom}_{\mathcal{W}}(X, X)$: for a $f \in \text{Hom}_{\mathcal{W}}(X, Y)$, we have $f \circ \text{id}_{X}=f, \text{id}_{Y}\circ f = f$; \vspace{1.5mm}
    \item Given two categories $\mathcal{W}$ and $\mathcal{W}^{'}$, we define the functor $F$: $\mathcal{W} \rightarrow \mathcal{W}^{'}$, which is similar to the function and maps objects with $F$: \text{Ob}($\mathcal{W}$) $\rightarrow$ \text{Ob}($\mathcal{W}^{'}$) and morphisms with $F$: $\text{Hom}_\mathcal{W}(X, Y) \rightarrow \text{Hom}_\mathcal{W}^{'}(F(X), F(Y))$;
    \vspace{1.5mm}
    \item Given a category $\mathcal{W}$, we define its opposite category $\mathcal{W}^{\text{op}}$: $\text{Ob}(\mathcal{W}^{\text{op}}) = \text{Ob}(\mathcal{W})$, $ \text{Hom}_{\mathcal{W}^{\text{op}}}(Y, X) = \text{Hom}_{\mathcal{W}}(X, Y)$;
    \vspace{1.5mm}
    \item Given categories $\mathcal{W}$ and $\mathcal{W}^{'}$, we define the isomorphism of functor $ F \simeq G $: if $ F:  \mathcal{W} \rightarrow \mathcal{W}^{'} $, $ G:  \mathcal{W}^{'} \rightarrow \mathcal{W} $, and $ F \circ G = \text{id}_{\mathcal{W}^{'}} $, $ G \circ F = \text{id}_{\mathcal{W}} $;
    \vspace{1.5mm}
    \item Given two categories $\mathcal{W}$ and $\mathcal{W}^{'}$, we define the isomorphism of categories $ \mathcal{W} \simeq \mathcal{W}^{'} $: if there are functors $ F:  \mathcal{W} \rightarrow \mathcal{W}^{'} $ and $ G:  \mathcal{W}^{'} \rightarrow \mathcal{W} $, $ F \circ G = \text{id}_{\mathcal{W}^{'}} $ and  $ G \circ F = \text{id}_{\mathcal{W}} $;
    \vspace{1.5mm}
    \item The category of sets, denoted as \textbf{Set}, is the category whose objects are sets. And the morphisms between its objects are the total functions. Naturally, the composition of morphisms in \textbf{Set} is the composition of functions;
    \vspace{1.5mm}
    \item  Given a category $\mathcal{W}$, a presheaf on the category $\mathcal{W}$ is a functor $F$: $\mathcal{W}^{op} \rightarrow $ $\textbf{Set}$;
    \vspace{1.5mm}
    \item Given a category $\mathcal{W}$, its category of presheaves is the functor category $ \mathcal{W}^{ \wedge } $ whose object is a presheaf functor $F$: $\mathcal{W}^{op} \rightarrow $ $\textbf{Set}$,
    which can be used to define the lifting functor that maps an object to its relationship;
    \vspace{1.5mm}
    \item Given any $X\in \text{Ob}(\mathcal{W})$, we define the Yoneda embedding $H_\mathcal{W}(X)$: $H_\mathcal{W}(X) \triangleq \text{Hom}_{\mathcal{W}}(\cdot, X)$. It provides a powerful and elegant way to represent objects in a category through functors: for any category, the objects within that category can be represented by presheaves;
    \vspace{1.5mm}
    \item 
    \wh{ We define $ K: \mathcal{W}^{\wedge} \rightarrow \mathbf{Set} $, a morphism in $ \mathcal{W}^{\wedge} $ that satisfies $K(H_{\mathcal{W}}(X),H_{\mathcal{W}}(Y)) = \text{Hom}_{\mathcal{W}}(X,Y)$. }
    \vspace{1.5mm}
    \item A task $T$ is a functor in $\mathcal{W}^{\wedge}$. Coding is also regarded as a specific task that only maintains the necessary information mentioned later.
\end{itemize}

\renewcommand{\baselinestretch}{1}


\vspace{1mm}

\noindent \textbf{Axioms}.
We list the following three Axioms as the basis of our research.
\begin{axiom}
\label{ax:rl}
[\textbf{Existence of Ideal Coding}] The relationship of ${X}$ captures all the information of ${X}$. (\textit{Analogy} Requirement)
\end{axiom}


This axiom is from the famous \textit{Yoneda lemma}~\cite{riehl2017category}, a fundamental lemma in category theory.
For $A \in \mathcal{W}^{\wedge}$, and ${X} \in \mathcal{W}$:
\begin{align}
  {\text{Hom}}_{\mathcal{W}^{\wedge}} \left( H_{\mathcal{W}}({X}), A \right) \simeq A({X}).
\end{align}

In category theory, we consider isomorphic relationships as being equal. Therefore, we can represent the relationship between $\mathcal{W}^{\wedge}$ and $\mathcal{W}$ in the following form: 
\begin{align}
  & \text{Hom}_{\mathcal{W}^{\wedge}} \left( H_{\mathcal{W}}(X), H_{\mathcal{W}}(Y) \right) \simeq  \nonumber
  \\
  & H_{W}(Y)(X) =  \text{Hom}_{\mathcal{W}}(X, Y).
\end{align}
This lemma illustrates that 
%
\wh{
\textit{\textbf{coding objects}} ($\text{Hom}_{\mathcal{W}}$) \textit{\textbf{equals coding their relationship}} ($\text{Hom}_{\mathcal{W}^{\wedge}}$).
\textit{\textbf{This relationship can be defined as intrinsic semantics.}}}
\vspace{0.5mm}

\begin{axiom}
\label{ax:mdl}
[\textbf{Existence of Practical Coding}] A functor $C$ in $\mathcal{W}^{\wedge} $ is an effective coding for a task functor $F$ if $F(X) \simeq \text{Hom}_{\mathcal{W}^{\wedge}} (C(X), F)$. (\textit{Abstraction} Requirement)
\end{axiom}

The process of coding is a process of information compression without introducing any new information.
$C$ is an effective coding means that, $C$ does not change any information from the perspective of $F$.
Therefore, we have:
\begin{align}
{\text{Hom}}_{\mathcal{W}^{\wedge}} \left( H_{\mathcal{W}}({X}),  F \right)  \simeq  {\text{Hom}}_{\mathcal{W}^{\wedge}} \left( C(X), F \right),
\end{align}
Thus, we can derive:
\begin{align}
F(X) \simeq {\text{Hom}}_{\mathcal{W}^{\wedge}} \left(H_{\mathcal{W}}({X}), F \right)  \simeq  {\text{Hom}}_{\mathcal{W}^{\wedge}} \left( C(X), F \right).
\end{align}
\textit{\textbf{This axiom, termed Coding Equivalence, indicates the essence of coding: reducing irrelevant redundant relationships from the target.}}

$C(X)$ can be a morphism to capture the real observed or collected relationship, 
\textit{e.g.} certain datasets or tasks, which capture the relationship in some dimensions of $H_{\mathcal{W}}(X)$.
The introduction of $C(X)$ is meaningful.
This Axiom is the natural extension of Axiom 1.
$H_{\mathcal{W}}({X})$ is not achievable or only obtained at a cost of infinite resources.
This axiom ensures the significance of coding in imperfect conditions and provides a guarantee:
\textit{\textbf{given any modality/task, the way to code them exists.}}
\vspace{0.5mm}

%

%

%
%
\begin{axiom}
\label{ax:mdl}
[\textbf{Compactness Leading to Generalization}]  It improves the generalization of coding when adding the compact constraint, \textit{e.g.} the shortest path. (\textit{Generalization} Requirement)
\end{axiom}

\wh{
}


We employ $MDL\left( \cdot \right)$, the minimal description length~\cite{rissanen1978mdlorigin,diederik2024aevb}
as a general measure or constraint of compactness and complexity.
For coding multiple objects $\{ X_i \}$, we have:
\begin{equation}
\label{eq:mdl2}
\footnotesize
MDL\left( \left\{ C(  X_i ), i=1,..., N \right\} | F \right) \leq MDL\left( C | F \right) +  \sum_{i=1}^{N} MDL\left( X_i | C, F \right).
\end{equation}

%
%

As $N$ is large, optimizing the right side will promote the decorrelation of $C$ and $X_i$ as well as different $X_i$ from the perspective of $F$.
Then, the information related to downstream tasks $F$ will be stored in $C$, and the corresponding representation of $X_i | {C}$ in the MDL view only maintains the randomness information that cannot be further compressed.
As demonstrated in existing theoretic explorations~\cite{blum2003lkm,diederik2024aevb,sefidgaran2023minimum}, their perfect decoupling capacity allows learning the invariance of intrinsic laws and the equivariance of stochastic factors, thus maintaining the generalization for downstream tasks.


\vspace{0.5mm}


\begin{figure}[t]
	\centering
	\includegraphics[width=0.4\textwidth]{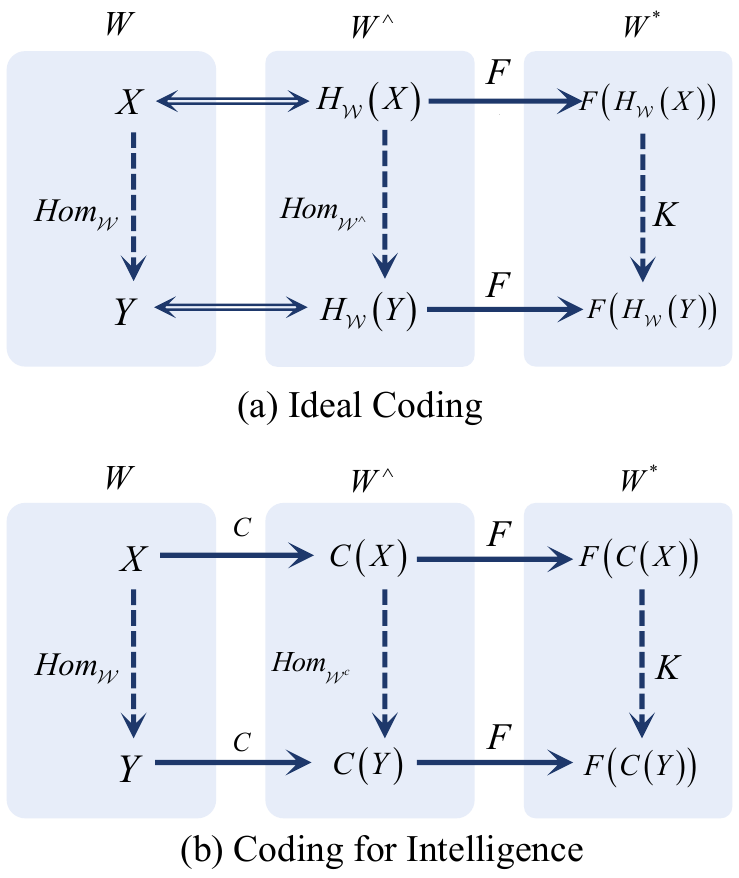}
	\caption{
        The definition of 
        (a) \textit{Ideal Coding}: using $K$ to approximate $\rm{Hom}_\mathcal{W}$ for coding $X$' intrinsic relationship $H_\mathcal{W}(X)$; 
        (b) \textit{Coding for Intelligence}: $C(X)$ captures only $X$'s relationship that are only meaningful to $F$, under the constraint of MDL.
	}
	\label{fig:cfi1}
\end{figure}

\wh{
These Axioms are closely associated with cognitive psychology, ensuring three points: 
1) Coding objects can be achieved by coding their relationship, which can be regarded as intrinsic semantics;
2) For any semantics/tasks, the way to code exists. 
3) Among all coding routes, there is an optimal way.
All these Axioms not only help define a novel CfI problem in the next subsection, 
but also provides support for the technical route of using engineering techniques to pursue the optimal coding solution.}

\subsection{Definition}
We make the following definitions:
\vspace{1mm}

\textit{\textbf{Definition 1}}: \textit{\textbf{Ideal Coding}}
aims to code intrinsic semantic relationships $H_{\mathcal{W}}(X)$ centered on the object $X$ in the category $\mathcal{W}$,
where $X$ is isomorphic with the \textit{intrinsic semantics} morphism $H_{\mathcal{W}}(X) \in W^{\wedge}$.
Then, $H_{\mathcal{W}}(X)$ is projected to the category $W^{*}$ with a task functor $F$ that acquires a subset of relationship in $H_{\mathcal{W}}(X)$.
The relationship of $X$, $Y \in \mathcal{W}$ can be easily determined by $F(H_{\mathcal{W}}(X))$ and $F(H_{\mathcal{W}}(Y))$ easily. (See Fig. 1 (a))
\vspace{1mm}

\textit{\textbf{Definition 2}}: \textit{\textbf{Coding for Intelligence (CfI)}}
aims to code observable semantics relationships $C(X)$ centered on the object $X$ in the category $\mathcal{W}$, with a coding lifting operator $C$ to a category $\mathcal{W}^{\wedge}$.
%
Then, $C(X)$ is projected to the category with a task functor $F$, 
where the relationship of $X$, $Y \in \mathcal{W}$ can be determined by $F ( C (X) )$ and $F ( C (Y) )$ easily. (See Fig. 1 (b))

CfI is the formal definition of practical coding mentioned in Section 2.2, which is achievable and can be taken as the target in the technical optimization stage.
We will clarify the meanings of these terms from the computer vision or image processing way.
$\mathcal{W}$ is an intrinsic scene or the world, and $X$ and $Y$ are specific objects within it. 
$H_{\mathcal{W}}(X)$ codes the ideal complete semantics of objects.
$C(X)$ and $C(Y)$ only capture parts of the semantics of objects, which are regarded as coded features.
%
$F(\cdot)$ can be considered as the prediction head of specific tasks that map the feature to the prediction result. 
$F(C(X))$ and $F(C(Y))$ are prediction results as either reconstruction or analytics results, 
$\mathcal{W}^{*}$ is the collection of prediction results.
It is noted that, $K$, which describes the intrinsic relationship of different objects in $\mathcal{W}^{*}$, 
plays not only the role of task definition but also that of training constraints/measures.
%
In the specific instances of CfI, the specific designs of $K$ lead to diverse but highly related tasks and 
$K$ is implemented as training losses.
Obviously, our approach can also model multiple tasks, simply by extending the formation of $F(C(\cdot))$, $\mathcal{W}^{*}$, and $K$ with multiple subscripts.

\subsection{Hint: How Coding Facilitates Intelligence}
From the psychological basis, axiom, and definition in the category theory view, we can obviously see the effect of coding for intelligence:
\begin{itemize}
    \item \textbf{Coding Relationship}.
    Coding data intrinsically is to compress its all relationships.
    \item \textbf{Coding as Information Bottleneck}.
    $C(X)$ \wh{filters} out useless redundancy in practice, improving the effectiveness of the system.
    \item \textbf{Coding as Regularizer for Generalization}.
    MDL perspective provides the optimizing direction of adopting the coding tools for improving the performance of whole intelligent systems. 
\end{itemize}

\section{Instances}
\label{sec:ins}

\begin{figure}[t]
	\centering
	\includegraphics[width=0.4\textwidth]{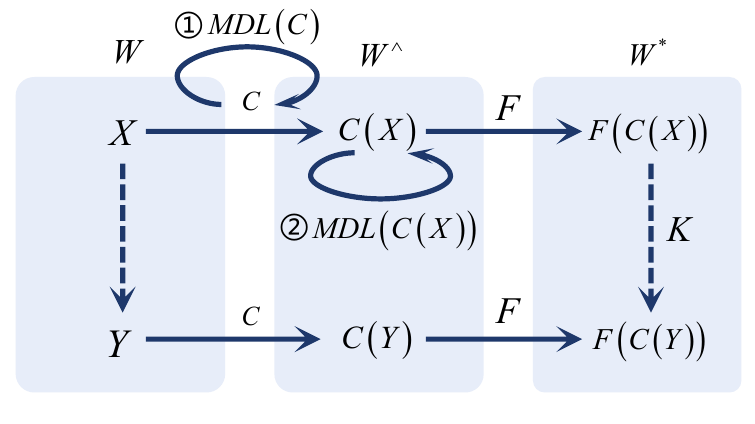}
	\caption{
        Specific instances under Coding for Intelligence.
	}
	\label{fig:ist}
\end{figure}

The specific instances are illustrated in Fig.~\ref{fig:ist}.
On the one hand, we hope to code $X$, \textit{i.e.} relationships of $X$ as much as possible, under the compact constraint.
On the other hand, the coding lifted $C(X)$ is expected to achieve desirable performance on diverse tasks
(not show the index of $i$ in Fig.~\ref{fig:ist} for simplicity).

Based on whether to optimize/constrain the compactness and complexity
in \textcircled{1} and \textcircled{2} of Fig.~\ref{fig:ist} and the form of $K$ in \textcircled{3},
we can obtain specific problem instances.

\subsection{Instances of $K$}
We discuss instances of $K$, which plays a role of the loss function for modeling diverse tasks.

\noindent \textbf{1) Reconstruction}. 
\vspace{0.7mm}

\noindent \textbf{Image Reconstruction}: not taking \textcircled{1} and \textcircled{2} as constraint and $K$ is defined as:
\begin{align}
\label{eq:rect}
\footnotesize
{K}\left( {{F_{{i}}}\left( {{C }\left( X \right)} \right),{F_{{i}}}\left( {{C }\left( Y \right)} \right)} \right) = \left\{ {\begin{array}{*{20}{c}}
{1,{\text{ if }} D(X, Y) \leq \epsilon, }  \\ 
{0,{\text{ if }} D(X, Y) > \epsilon. }
\end{array}} \right.
\end{align}
\wh{$D(\cdot, \cdot)$ measures the difference between $X$ and $Y$, and $\epsilon$ is small positive quantity.
There are many choices for $D(\cdot, \cdot)$ in practice, \textit{e.g.} L2 to measure the pixel distance, and Fr$\Acute{e}$chet Inception Distance (FID)~\cite{heuse2017fidl} for measuring the perceptual feature distance.
}
\vspace{0.7mm}

\noindent \textbf{Image Compression}: taking \textcircled{2} as the constraint and $K$ is defined the same as Eqn.~\eqref{eq:rect}.
\vspace{0.7mm}

\noindent \noindent \textbf{Image Restoration}: taking \textcircled{1} as the constraint and $K$ is defined as follows,
\wh{
\begin{align}
\label{eq:ir}
\footnotesize
& {K}\left( {{F_{{i}}}\left( {{C }\left( M(X) \right)} \right),{F_{{i}}}\left( {{C }\left( M(Y) \right)} \right)} \right)  \nonumber  \\
 = & 
 \left\{ {\begin{array}{*{20}{c}}
{1,{\text{ if }} D(M(X), M(Y)) \leq \epsilon, } 
\\
{0,{\text{ if }} D(M(X), M(Y)) > \epsilon. }
\end{array}} \right.
\end{align}
where $M$ is a degradation operation.
}

\begin{figure*}[t]
	\centering
	\includegraphics[width=\textwidth]{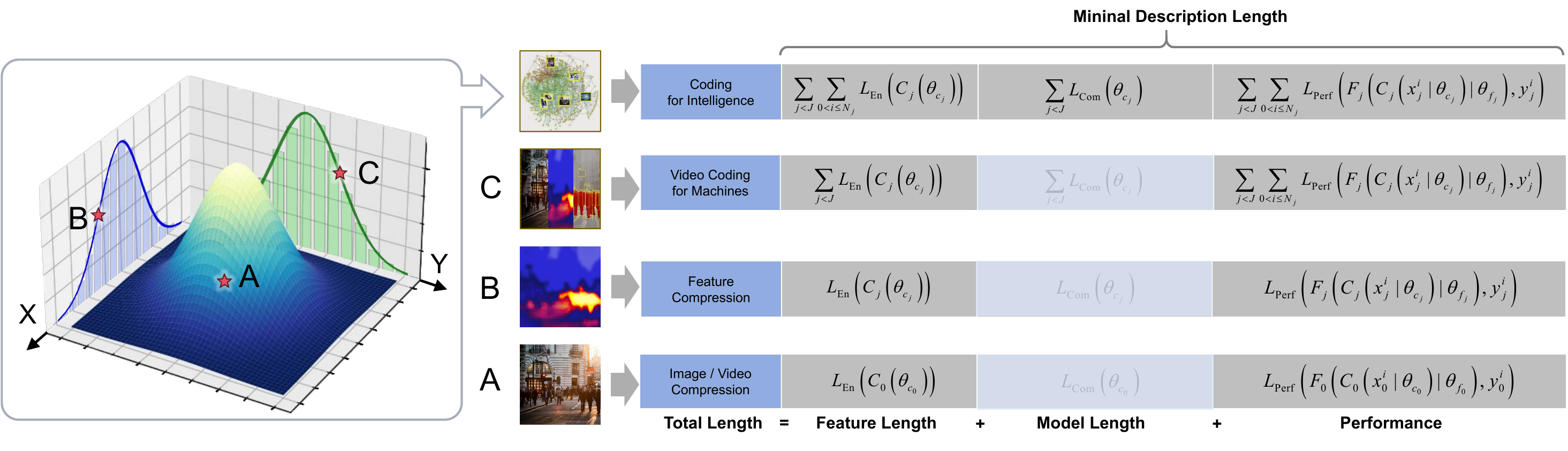}
	\caption{
        Specified MDL formulation for different CfI instances.
        Image coding (A) and feature coding (B) aim to build a compact representation for given samples.
        Coding for machines (C) hopes to consider compactness from the perspective of the joint probability distribution.
        Coding for intelligence (D) expects to compress the whole thing from both perspectives of sample compactness and joint distribution compactness.
	}
	\label{fig:mdl}
\end{figure*}

\vspace{0.7mm}
\noindent \textbf{2) Classification}
\vspace{0.7mm}

\noindent \textbf{Supervised Learning}: taking \textcircled{1} as the constraint and $K$ is defined as follows,
\begin{align}
\label{eq:class}
\footnotesize
{K}\left( {{F_{{i}}}\left( {{C }\left( X \right)} \right),{F_{{i}}}\left( {{C }\left( Y \right)} \right)} \right) = \left\{ {\begin{array}{*{20}{c}}
{1,{\text{$X$, $Y$: Same Category}},}\\
{0,{\text{$X$, $Y$: Different Category}}.} 
\end{array}} \right.
\end{align}

\noindent \textbf{Feature Compression Based on Fully Supervised Learning}: taking \textcircled{2} as the constraint and $K$ is defined the same as Eqn.~\eqref{eq:class}.

\vspace{0.7mm}
\noindent \textbf{3) Contrastive Learning}
\vspace{0.7mm}

\noindent \textbf{Unsupervised Learning}: taking \textcircled{1} as the constraint and $K$ is defined as follows,
\begin{align}
\label{eq:contrast}
\footnotesize
{K}\left( {{F_{{i}}}\left( {{C }\left( X \right)} \right),{F_{{i}}}\left( {{C }\left( Y \right)} \right)} \right) = \left\{ {\begin{array}{*{20}{c}}
{1,{\text{ $Y = T(X)$ }},}\\
{0,{\text{ $Y \ne T(X)$ }},} 
\end{array}} \right.
\end{align}
where $T$ is the augmentation transform defined in contrastive learning.

\noindent \textbf{Feature Compression based on Contrastive Learning}: taking \textcircled{2} as the constraint and $K$ is defined the same as Eqn.~\eqref{eq:contrast}.

\vspace{1mm}
\noindent \textbf{4) Mask Image Modeling}
\vspace{1mm}

\noindent \textbf{Unsupervised Learning}: taking \textcircled{1} as the constraint and $K$ is defined as follows,
\begin{align}
\label{eq:mim}
\footnotesize
{K}\left( {{F_{{i}}}\left( {{C }\left( X \right)} \right),{F_{{i}}}\left( {{C }\left( Y \right)} \right)} \right) = \left\{ {\begin{array}{*{20}{c}}
{1,{\text{ $Y = M(X)$ }},}\\
{0,{\text{ $Y \ne M(X)$ }},} 
\end{array}} \right.
\end{align}
where $M$ is the mask operation.
\vspace{1mm}

\noindent \textbf{Summary}:
1) Our definition is general for covering all computational-related topics towards AI.
Therefore, with the connection built with category, we can effectively describe a wide range of AI computational problems while also smoothly shifting to more concrete forms;
2) $K$ is an ideal tool for describing relationships but may not be a feasible option. Metrics/losses need to be designed to describe or simulate $K$ for model design and training;
3) While category theory reveals the homogeneity of these methods, how to achieve a more unified optimization on a larger scale needs more discussion.
Therefore, in the following, we introduce a specific form of MDL that leads to revealing potential routes to intelligence.

\subsection{Instances to Model $\rm{Hom}$ with the Guidance of $K$}
To effectively \wh{capture and approximate} $\rm{Hom}$, the learning models continuously evolve.
Despite there are infinite ways the information flows within the model, and there can theoretically be an infinite number of model patterns,
from the perspective of successful practice, the existing advanced models primarily have two ways to capture and organize information.
\vspace{1mm}

\noindent 
\wh{
\textbf{1) Progressive chain-like structures}. 
To model the relationship between $X$ and $Y$, one technical route is to sample a logical chain in $\mathcal{W}$ space: \(X \rightarrow E_1 \rightarrow E_2 \rightarrow \ldots \rightarrow Y\).
The morphism $\text{Hom}_{\mathcal{W}}(X,Y) $ can be expressed as
$ \text{Hom}_{\mathcal{W}}(X, E_1) \circ \text{Hom}_{\mathcal{W}}(E_1, E_2) \circ \ldots \circ \text{Hom}_{\mathcal{W}}(E_n,Y) $. 
This modeling pattern is used by Recurrent Neural Networks (RNN)~\cite{hu2018pro,hochreiter1997lstm}, Large Language Models (LLM)~\cite{touvron2023llama}, and diffusion models~\cite{ho2020denoising}.
RNNs process sequences by looking only at the information from the last time-step at each stage, making it effective for tasks involving temporal dependencies.
For LLMs that apply next token prediction,  $E_i$ is taken as the prediction target of $E_{i-1}$.
For diffusion models, $X$ is Gaussian noise and $Y$ is an image, and $E_i$ is the noisy image.
Adding a chain in front of $X$ is similar to the mechanism of applying prompt~\cite{huang2023densityprompt,liu2023explicit} and retrieval-augmented generation (RAG)~\cite{gao2023rag}.
Ideally, adding a chain between $X$ and $Y$ corresponds to the chain of thoughts.
}
%
%
%
%
\vspace{1mm}

\noindent \textbf{
2) Structural abstraction and reparameterization}, using compact and efficient parameters to form gradually abstract information for manipulating the modeling process.
\wh{
In this case, logical chain sampling is performed in $ \mathcal{W}^{\wedge} $  space:
\begin{align}
    H_{\mathcal{W}}(X) \rightarrow S_1 \rightarrow \ldots \rightarrow S_n, \nonumber \\
    H_{\mathcal{W}}(Y) \rightarrow S_1 \rightarrow \ldots \rightarrow S_n. \nonumber
\end{align}
Then, the relationship at each level is sampled/inferred based on the last level's relationship. 
This modeling process can be expressed as follows,
\begin{align}
& \text{Hom}_{\mathcal{W}}(X,Y)  = \text{Hom}_{\mathcal{W}^{\wedge}}(\text{H}_{\mathcal{W}}(X), \text{H}_{\mathcal{W}}(Y)) \approx \nonumber  \\
& \text{Hom}_{\mathcal{W}^{\wedge}}(\text{H}_{\mathcal{W}}(X), S_1) \circ \text{Hom}_{\mathcal{W}^{\wedge}}(S_1, S_2) \circ \ldots \circ  \nonumber \\
& \text{Hom}_{\mathcal{W}^{\wedge}}(S_{k-1}, S_k) \circ \text{Hom}_{\mathcal{W}^{\wedge}}(S_{k}, S_{k-1}) \circ \ldots \circ \nonumber \\
& \text{Hom}_{\mathcal{W}^{\wedge}}(S_2, S_1) \circ \text{Hom}_{\mathcal{W}^{\wedge}}(S_1, \text{H}_{\mathcal{W}}(Y)), \nonumber
\end{align}
For example, in CNN models~\cite{alex2017alexnet, he2016resnet}, features at a given level are obtained from the last one through convolutions, 
thus the relationship at each level aggregates from the relationships of previous levels, corresponding to the transition from local to global scales. 
Each $S_i$ can be considered as sampling for a task, thus different layers in CNNs can be used for different tasks. 
Therefore, the relationship between $X$ and $Y$ turns to the relationship of $X$ and $S$'s relationship and $Y$ and $S$'s relationship. 
%
%
Variational auto-encoder (VAE)~\cite{kingma2019vae} and normalizing flow~\cite{kobyzev2021normal} are also important works in this category.
VAE learns to encode input data into a latent space and then decodes it back to the original space.
This learning process can explicitly reparameterize the network representation into a compact data distribution representation.
A recent progress is the normalizing flow~\cite{kobyzev2021normal}. 
It can transform a simple probability distribution into a more complex one through 
a series of invertible and differentiable mappings, enabling flexible and efficient density estimation and data generation.
}

\wh{
In Section~\ref{sec:trend}, it is observed that, in the modeling and compression of multi-modality data for multiple tasks, significant coding pathways are emerging along these two directions.
In Section~\ref{sec:diss}, we can also see that, these two paradigms of modeling $\text{Hom}$ also met in the large foundation model realm.
}

\subsection{Specification: Minimal Depth Length Paradigm}
In this subsection, we further implement the formulation in the form of category theory to a specific linearly combined loss constraint,
which is more friendly for computation and optimization.
\begin{align}
\label{eq:vcm}
U & =  \sum\limits_{j < J}^{} {\sum\limits_{0 < i \le {N_j}}^{} {{L_{{\rm{Perf}}}}\left( {{F_j}\left( {{C_j}\left( {x_j^i|{\theta _{{c_j}}}} \right)|{\theta _{{f_j}}}} \right),y_j^i} \right)} } \nonumber \\
&  + \sum\limits_{j < J}^{} { \sum\limits_{0 < i \le {N_j}}^{} {{L_{{\rm{En}}}}\left( {{C_j}\left( {x_j^i|{\theta _{{c_j}}}} \right)} \right)} }  + \sum\limits_{j < J}^{} {{L_{{\rm{Com}}}}\left( {{\theta _{{c_j}}}} \right)} \nonumber \\
&  + \sum\limits_{j < J}^{} {{L_{{\rm{Com}}}}\left( {{\theta _{{f_j}}}} \right)},   
\end{align}
where $x$ is the input data, $j$ indexes the task and $i$ indexes the sample for $j$-th task.
For simplicity, we define that $x_j$ is the pixel signal, and $x_j (j>0)$ is the semantic feature that corresponds to diverse semantic tasks.
We define that $C_j (\cdot| \theta_{c_j} )$ and $F_j (\cdot| \theta_{f_j} )$ are encoder/compressor and decoder/decompressor,
where $\theta_{c_j}$ and $\theta_{f_j}$ are their parameters to be optimized, respectively.
Note that, although $C$ is expected to be as general as possible and not depend on specific tasks, in the practical implementation,
existing frameworks usually still take $C$ as differentiated ones.
$J$ is the task number and $N_j$ is the sample number in the task $j$.
For convenience, on the symbolic level, we do not make detailed distinctions between the compression and (downstream) analytics models.
Loss terms are explained in the following:
\begin{itemize}
    \item $L_{{\rm{Perf}}}(\cdot)$ represents the task performance, which describes the fitness of relationship $K$;
    \item $L_{{\rm{En}}}(\cdot)$ measures the entropy of the representation, \textit{i.e.} the length of a signal given the coding knowledge, 
    corresponding to the third term in Eqn.~\eqref{eq:mdl2}.
    \item $L_{{\rm{Com}}}(\cdot)$ calculate the parameters' complexity, corresponding to the first and second terms in Eqn.~\eqref{eq:mdl2}.
\end{itemize}
%


From a historical perspective, these two areas started with distinct trajectories and then gradually fused.
The Turing Test was proposed in 1950, which is taken as the gold standard for evaluating whether an AI is close to human intelligence or not.
Later, in 1956, the term AI was coined~\cite{kaplan2019siri} in a workshop held on the campus of Dartmouth College.
The perception is proposed in 1969, which gradually evolves into neural networks, laying the foundation for today's deep learning frameworks.
%
However, since 1970, both in academia and the industry, enthusiasm for AI has gradually faded, leading to AI winter.
In nearly 2000, AI once again achieved a breakthrough, accompanied by a series of new highlights, 
including deep blue~\cite{campbell2002deepblue} (the first chess program that beat human champion), 
Kismet~\cite{Kismet} (can recognize and simulate emotions), \textit{etc}.
\wh{These evolutions attribute to optimizations for $L_{{\rm{Perf}}}(\cdot)$ while constraining $L_{{\rm{Com}}}(\cdot)$ to a certain level.}

In another branch, in the field of coding, as early as 1927, Ray Davis Kell registered a patent that was remarkably ahead of its time, containing the content about transmitting successive complete images of the transmitted picture.
From the theory perspective, the efforts of Hartley~\cite{Har28} and Shannon~\cite{shannon} established the mathematical foundations and built the modern information theory, which heavily affects the development of coding techniques.
Since then, at various stages of coding technology development \wh{to reduce the cost of $L_{{\rm{En}}}(\cdot)$}, it has been gradually propelled and advanced by AI models.
The earliest methods directly apply \textit{entropy coding}, \textit{e.g.} Huffman coding~\cite{huffman}, and arithmetic coding~\cite{witten1987arithmetic}, to reduce statistical redundancy.
In later 1960, the \textit{transform} that makes image energy compact, \textit{e.g.} Fourier transform~\cite{pohlig1980ft} 
and Discrete Cosine Transform (DCT)~\cite{ahmed1974dct}, is introduced to make the representation more compact.
Later works, \textit{e.g.} JPEG standard~\cite{wallace1992jpeg}, incorporate \textit{prediction function} into the coding model, which use the prediction of elements of the current signal to remove the redundancy.
This trend also gives birth to the hybrid transform/prediction coding paradigm in video coding. The temporal redundancy can be removed by inter-frame prediction.
Under this framework, many of these classical video coding standards, \textit{e.g.} H.264/AVC~\cite{kalva2006h264}, AVS (Audio and Video coding Standard)~\cite{fan2004avs}, and HEVC~\cite{sullivan2012hevc}, have been established.

In the past ten years, deep learning powers and its flexible modeling mechanisms have led to the fusion of two kinds of categories.
Since 2016, deep neural networks~\cite{park2016inloop,liu2016cupar,li2018line,xia2018group,dai2018cuclass} become part of modules in conventional image/video codecs.
Meanwhile, the end-to-end learnable frameworks are also developed, including generalized divisive normalization (GDN)~\cite{johannes2016gdn,balle2017endtoend}, recurrent neural network (RNN)~\cite{toderici2017rnn}, variational autoencoder~\cite{diederik2024aevb}, \textit{etc.}
These methods have gradually surpassed the performance of traditional codecs over the years and demonstrated more flexible characteristics.
In the other direction, entropy serves as a fundamental element in model construction, driving AI models to achieve performance leaps, \textit{e.g.} cross-entropy loss~\cite{cover2006info}, information-theoretic loss~\cite{xu2019infotheory}, and K-L divergence~\cite{csiszar1975divergence}, \textit{et al}. 
Besides, the module design of bottleneck architecture~\cite{lin2013network,Ronneberger2015unet}, dropout~\cite{srivastava2014dropout}, pooling~\cite{he2015spp}, mask modeling~\cite{he2022mae}, points out the positive effect of compact representations in screening critical semantic information while filtering out noise.
Some works reveal that deep learning is a bottleneck mechanism~\cite{michael2018on,naftali2015ibp}.
In recent works~\cite{icml2023kzxinfodl,yu2020learning}, the entropy constraint is embedded as an effective regularizer to augment the feature's capacity in the general sense. 
\vspace{0.7mm}

\textbf{Summary}: As evident from the formulas, the deep integration and development of these two fields today are inseparable from the fact that they stem from the same fundamental scientific problem of MDL.
We summarize their relationship from the MDL view in Fig.~\ref{fig:mdl}.
Image/feature compression targets to optimize the performance of image reconstruction ($0$-th task) / analytics ($j$-th task) under the constraint of bitrate jointly.
Comparatively, video coding for machines considers the R-D cost under the multi-task scenario, which implicitly models the task relationship.
\wh{CfI defined in this paper further} considers from a more comprehensive view to form the joint optimization taking the inputs from different domains for multiple tasks.

\subsection{Discussion: from Relationship $K$ to Distance $L$}
%
%

$K$ defines the relationship in the ideal case. However, in the real world, 
observing and measuring this relationship in model construction and inference is often challenging. Therefore, in practice, 
we need to design a specific loss function $L$ to describe and constrain the object relationship.
Based on the nature of the constraint, we briefly discuss the design of loss functions from two categories:
supervised learning and unsupervised learning.
The former describes the one-to-one relationship between the input and its corresponding ground truth,
obtaining the intrinsic semantics from the input given a specific task context.
It takes the form of measuring the distance between the prediction and ground truth.
The latter describes the soft semantic relationship among different objects in a category.
It adopts the form of measuring the similarity of those objects.
\vspace{2mm}

\noindent \textbf{Fully Supervised Learning.} 
Fully supervised learning aims to capture the mapping relationship described in $K$, usually provided from the collected labeled data.
Generally, it can be further classified into two categories: regression-based and recognition-based.
The former focuses on regressing the signal precisely and is usually widely adopted in reconstruction-based tasks.
The commonly used ones include 
fidelity-driven losses, \textit{e.g.} $L_1$ loss, $L_2$ loss, smooth $L_1$ loss~\cite{lai2019lpn},
structure-driven losses, \textit{e.g.} Structural SIMilarity (SSIM)~\cite{wang2004ssim}, 
and perception-driven losses, \textit{e.g.} Learned Perceptual Image Patch Similarity (LPIPS)~\cite{zhang2018lpips}, Deep Image Structure and Texture Similarity (DISTS)~\cite{ding2022dist}.
The latter aims to measure the feature distance according to the intrinsic semantic space, determined by the downstream tasks.
The widely used ones include
softmax loss~\cite{balcan2016softmax}, cross-entropy loss~\cite{mao2023cross_entropy}, Hinge loss~\cite{luo2021shl} for image classification, 
intersection-over-union (IoU)~\cite{zhou2019iou}, focal loss~\cite{lin2017focal} for object detection or semantic segmentation, \textit{etc}.
These aspects also see emerging new work.
The main focus is still on addressing two issues: 
1) The sampled label data is biased, hence the desire for the metrics themselves to be robust enough;
2) There is a gap between the categories of $L$ and $K$, so there is still room for improvement in the approximation of the measure.
\vspace{2mm}

\begin{table*}[!tbp]
        \footnotesize
	\caption{
		Summary of methods to code different media and their required storage, \textit{e.g.} descriptor size and bit rate.
	}
	\label{tab:objects}
	\vspace{-3mm}
	\centering
	\resizebox{\textwidth}{!}{
	\begin{tabular}{cccrrrc}
		\hline
		Objects                         &  Applications                              & Size    & Bitrate & Dataset & Reference &     \\
		\hline
		\hline
		Text (Lossless)   
        &  \cellcolor[HTML]{EFEFEF} Text-Related Tasks  &  -  & 0.7-1.5 (bpc) &  \cellcolor[HTML]{EFEFEF}
		\begin{tabular}[r]{@{}r@{}} 
        Text8~\cite{text8_dataset} 
        \\ 
        Gutenberg~\cite{gutenberg}
		\end{tabular}
		& Valmeekam \textit{et al.} 2023 & \cite{valmeekam2023llmzip} \\
		Audio  &  \cellcolor[HTML]{ECF4FF} Audio-Related Tasks  &  -  &  1.5-24 (kbps)  &  \cellcolor[HTML]{ECF4FF}
		\begin{tabular}[r]{@{}r@{}}
		AudioSet~\cite{gemmeke2017audio}
        \\ DAPS~\cite{mysore2015audio}
		\\ MUSDB~\cite{musdb18}
	    \end{tabular}
		& Kumar \textit{et al.} 2023 & \cite{kumar2023highfidelity} \\
		\hline
		\hline
		\begin{tabular}[l]{@{}c@{}} Compact Descriptors \\ for Visual Search (CDVS) \end{tabular}     & \cellcolor[HTML]{EFEFEF} Matching/Retrieval                        & 512B-16KB          & -                 &  \cellcolor[HTML]{EFEFEF}  \begin{tabular}[r]{@{}r@{}}CDVS~\cite{cdvs_dataset} \end{tabular} &  Duan \textit{et al.} 2015 & \cite{duan2015cdvs}   \\
		Deep Hashing       &  \cellcolor[HTML]{ECF4FF} Matching/Retrieval   & 4B-128B            & -    &  \cellcolor[HTML]{ECF4FF} \begin{tabular}[l]{@{}c@{}}
		\begin{tabular}[r]{@{}r@{}}
	    VehicleID~\cite{liu2016cvpr} \\
		VERI-Wild~\cite{bai2022veriwild2} \\
		Market1501~\cite{zheng2015market1501} \\
	    MSMT17~\cite{zheng2015market1501}
	    \end{tabular}
		\\
	    \end{tabular}
		&  Duan \textit{et al.} 2018 & \cite{duan2018tip}          \\
		Deep Intermediate Features          &  \cellcolor[HTML]{EFEFEF} Classification/Detection  & -     & 0.624-4.28 (bpp) & \cellcolor[HTML]{EFEFEF} ImageNet~\cite{deng2009imagenet}       &  Chen \textit{et al.} 2019  & \cite{chen2020intermediate}          \\
		Deep Learned Features                  & \cellcolor[HTML]{ECF4FF} Tasks Involved in Training                &    -    &  0.1-1.2 (bpp)  & \cellcolor[HTML]{ECF4FF}
		\begin{tabular}[l]{@{}c@{}}COCO2014~\cite{lin2014coco}
		\end{tabular} 
		&  Choi \textit{et al.} 2022 & \cite{choi2022scalable}  \\
		Semantic Maps (Lossless)               & \cellcolor[HTML]{EFEFEF} \begin{tabular}[l]{@{}c@{}} Pixel-Level \\ Semantics Understanding\end{tabular}   &  -   & 5.2*10-5-0.0776 (bpp)    &    \cellcolor[HTML]{EFEFEF}
		\begin{tabular}[l]{@{}c@{}} Cityscapes~\cite{cordts2016cityscapes} \end{tabular}
		& Yang \textit{et al.} 2020 & \cite{yang2020pcc}        \\
		Semantic Maps (Lossy)                  & \cellcolor[HTML]{ECF4FF} \begin{tabular}[l]{@{}c@{}}Pixel-Level \\ Semantics Understanding\end{tabular}       & -                  &    0.004-0.006 (bpp)           &    \cellcolor[HTML]{ECF4FF} Taskonomy~\cite{zamir2018taskonomy}  &  Yang \textit{et al.} 2024 & \cite{yang2024compresstaskonomy}         \\
		Point Sequence                         & \cellcolor[HTML]{EFEFEF} \begin{tabular}[l]{@{}c@{}}Action Recognition \\ and Analysis \end{tabular}    & 2.46B-94.36B       & - & \cellcolor[HTML]{EFEFEF}
		\begin{tabular}[l]{@{}c@{}}
		    HiEve~\cite{lin2020HumanIE}
		\end{tabular}
		&   Lin \textit{et al.} 2020 & \cite{lin2020pointcompress}        \\
		Shading                                & \cellcolor[HTML]{ECF4FF} \begin{tabular}[l]{@{}c@{}} 
			3D Modeling and \\ Scene Analysis\end{tabular} &     -     &    0.004-0.01 (bpp)    &  \cellcolor[HTML]{ECF4FF} Taskonomy~\cite{zamir2018taskonomy}        &  Yang \textit{et al.} 2024 & \cite{yang2024compresstaskonomy}  \\
		Surface Normal  
		&  \cellcolor[HTML]{EFEFEF} \begin{tabular}[l]{@{}c@{}} 
			3D Modeling and \\ Scene Analysis\end{tabular}
		&        -              &        0.007-0.028 (bpp)     &  \cellcolor[HTML]{EFEFEF}  Taskonomy~\cite{zamir2018taskonomy}    &   Yang \textit{et al.} 2024 & \cite{yang2024compresstaskonomy}        \\
		Scene Class                            &  \cellcolor[HTML]{ECF4FF} Classification                            &       -              &              0.002-0.006 (bpp) &  \cellcolor[HTML]{ECF4FF}  Taskonomy~\cite{zamir2018taskonomy}       &   Yang \textit{et al.} 2024 & \cite{yang2024compresstaskonomy}       \\
		\hline
		\hline
		RGB Image (Fidelity)                   &  \cellcolor[HTML]{EFEFEF} \begin{tabular}[l]{@{}c@{}} Reconstruction  \end{tabular}        &  -  &  0.1-1.2 (bpp)       & \cellcolor[HTML]{EFEFEF} CLIC~\cite{clic} &  Wang \textit{et al.} 2022 & \cite{wang2022neuraltrans} \\
		RGB Image (Perceptual)                 &  \cellcolor[HTML]{ECF4FF} \begin{tabular}[l]{@{}c@{}}Reconstruction \end{tabular}  &  -  &  0.1-0.4 (bpp) &  \cellcolor[HTML]{ECF4FF}  CLIC~\cite{clic}  &  Mentzer \textit{et al.} 2020 & \cite{mentzer2020high}  \\
		RGB Image (Conceptual)                 &  \cellcolor[HTML]{EFEFEF}  \begin{tabular}[l]{@{}c@{}}Reconstruction and \\ Downstream Tasks  \end{tabular}  &  -  &  0.02-0.15 (bpp) &  \cellcolor[HTML]{EFEFEF} FFHQ-Aging~\cite{orel2020lifespan}  &  Yang \textit{et al.} 2023 & \cite{yang2023manifold} \\		
		Point Cloud          &                 \cellcolor[HTML]{ECF4FF}     \begin{tabular}[l]{@{}c@{}}Reconstruction  \end{tabular}                                        &  -  & 1-8 (bpp) & \cellcolor[HTML]{ECF4FF} SemanticKITTI~\cite{behley2019SemanticKITTI} &  He \textit{et al.} 2022 & \cite{he2022density_pcc} \\
		\begin{tabular}[l]{@{}c@{}}
			Multispectral \\ Image
		\end{tabular}
		&  \cellcolor[HTML]{EFEFEF} \begin{tabular}[l]{@{}c@{}}Reconstruction \end{tabular}                                         &  -  &  0-2 (bpp) & 
        \cellcolor[HTML]{EFEFEF}
		\begin{tabular}[l]{@{}c@{}}
		AVIRIS~\cite{AVIRIS}
		\end{tabular}
		&  Meyer \textit{et al.} 2020 & \cite{Meyer2020Multispectral} \\
		Stereo Image                          &  \cellcolor[HTML]{ECF4FF}  \begin{tabular}[l]{@{}c@{}}Reconstruction \end{tabular}                                         &  -   &  0.1-0.7 (bpp) & \cellcolor[HTML]{ECF4FF} InStereo2K ~\cite{Bao2020InStereo2KAL} &  Wödlinger \textit{et al.} 2022 & \cite{Wodlinger2022SASIC} \\
		Medical Image                         &  \cellcolor[HTML]{EFEFEF} \begin{tabular}[l]{@{}c@{}}Reconstruction \end{tabular}                                         &  -  & 0.1-1.8 (bpp) & \cellcolor[HTML]{EFEFEF} NIH~\cite{nih} &  Guo \textit{et al.} 2023 & \cite{guo2023xray}  
		\\ 
		Event                         &  \cellcolor[HTML]{ECF4FF} \begin{tabular}[l]{@{}c@{}}
		Reconstruction \end{tabular}                                        &  -  & 50-2500 (cr) & \cellcolor[HTML]{ECF4FF} RGB-DAVIS~\cite{wang2020rgbdavis} 
		&  Banerjee \textit{et al.} 2021 & \cite{banerjee2021event}
		\\
		\hline
        \hline
	\end{tabular}}
\end{table*}

\begin{figure}[t]
	\centering
	\includegraphics[width=0.5\textwidth]{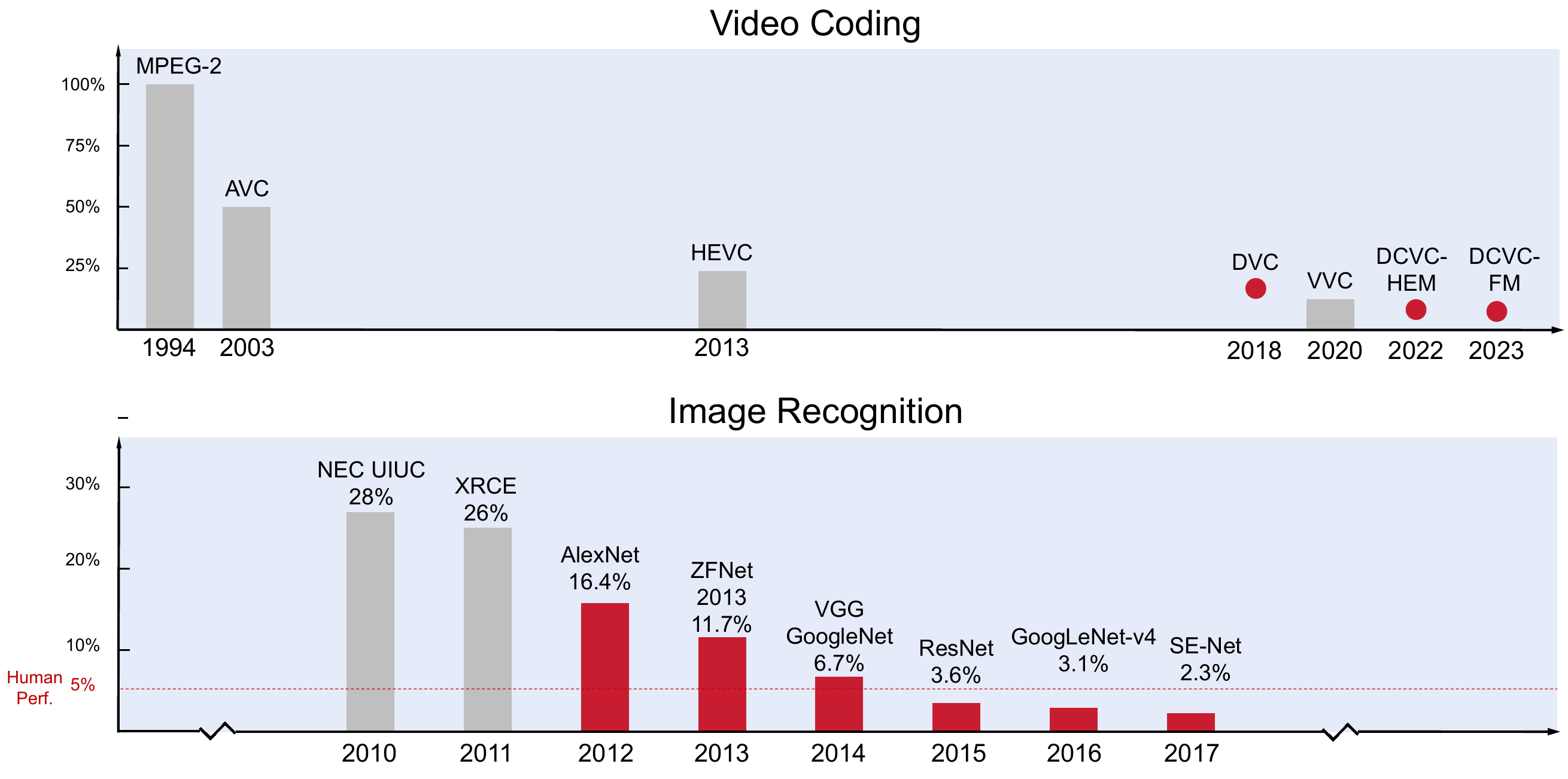}
	\caption{
        Deep learning has accelerated the development of AI and coding techniques,
        driving machine-driven image recognition beyond human cognitive capabilities and
        rapidly advancing end-to-end video encoding beyond traditional codecs.
	}
	\label{fig:his}
\end{figure}

\noindent \textbf{Unsupervised Learning.} 
The measures of unsupervised learning do not rely on labels, aiming to describe the sample's relationship.
In the early stages of the work, these methods are similar to those adopted for supervised losses, 
with the difference lying in pre-transforms applied before computing the final metrics, \textit{e.g.} clustering~\cite{nie2023kmean}, PCA~\cite{jolliffe2016pca}, \textit{etc}.
Recent generative models bring in new revolutions by using deep networks as the constraint, which shows the potential of learning $L$ to approach $K$.
A typical example comes from the generative adversarial network~\cite{goodfellow2014gan} using two loss functions.
The minimax loss~\cite{goodfellow2014gan} directly measures the cross-entropy between real and generated data distributions in the adversarial play.
The Wasserstein loss~\cite{martin2017icml} focuses on minimizing the distance between the distribution of the training and generated datasets.
There are also works on unsupervised/self-learning for recognition that enrich the power of losses with diverse transforms.
Masked autoencoders (MAE)~\cite{he2022mae} combines masking operation and signal fidelity-driven losses.
Contrastive learning-based methods ~\cite{chen2020simclr} aim to learn the general feature representation via the intrinsic paired similarity.
\vspace{2mm}

\noindent \textbf{Summary}: Through the efforts of two types of loss functions, we are increasingly capable of characterizing $K$, especially with the advent of learnable $L$ (\textit{e.g.}, adversarial learning, large language models), which brings significant gains.
Recent work~\cite{nowozin2016fgan,zhang2022fmutual,rosasco2004same} suggests that many loss functions are inherently related or interconnected. However, on the one hand, whether $L$ can effectively characterize $K$, and whether there exist cases where $K$ cannot be described in concrete forms or measured, remains an unexplored issue, which will be discussed in Section~\ref{sec:diss}.

\section{Summary of Previous Works}
\label{sec:sum}

In this section, we introduce the recently emerging instance under the CFI definition that handles data of multiple domains for multiple tasks as shown in Fig.~\ref{fig:cfi}.
Then, we turn the problem into a more specific linear multi-objective optimization function in Eqn.~\eqref{eq:vcm}.
Based on three critical elements, \textit{i.e.} data, model, task, we review the development of recent progress from three aspects.

\subsection{Diverse Coding Objects}

In the field of coding, images, and videos stand out as the earliest and most thoroughly studied objects.
However, the inherent signal forms in our world exhibit considerable diversity,
leading to a corresponding diversity in the objects that require coding, storage, and analysis.

To simplify, the diversity of coding objects arises from three perspectives: natural modality, capture device, and application requirement.
The first scenario is related to naturally divergent data form in the physics world, resulting in different signals and being represented as different types,
\textit{e.g.} image, video, sound, text, \textit{etc}.
Second, the diverse sensors in smart cities naturally also collect data of different forms, \textit{e.g.} point cloud, infrared images.
Third, another need comes from the application requirements of handling multiple tasks in intelligent analytics.
The intermediate processed features and final predicted results of various intelligent tasks are regarded as another source of coding modalities, \textit{e.g.} deep features or semantic maps.

We summarize the coding objects and their rough properties in Table~\ref{tab:objects}.
Examining these media sources, it becomes apparent that existing technologies have already conducted in-depth research on individual modality data.
When gathering these correlated features, they have the potential to contribute to constructing joint optimization for superior compression performance.
In addition, we see a diversification in their bitrates. 
The bitstream of text and audio has a small size. The bit-rates of images and videos are a bit relatively high.
The features that are used directly for specific downstream tasks,
\textit{e.g.} deep learned features, or highly correlated features, \textit{e.g.} shading, surface normal, also have a relatively small size.
Comparatively, the features for general visual analytics, \textit{e.g.} deep intermediate features, take high bit-rates.
Therefore, when addressing the challenge of jointly compressing multi-modality data, the critical necessity emerges for hierarchical feature organization in a scalable way.
These techniques facilitate obtaining an integrated compact representation for multiple tasks with a unified scalable bitstream, which triggers a series of novel technical routes recently.

%
%
%

\subsection{Evolving Models}

Artificial intelligence technology has undergone a series of ups and downs in its development in history.
The Turing Test was proposed in 1950,
which is taken as the gold standard for evaluating whether an AI is close to human intelligence or not.
Later, in 1956, the term AI was coined~\cite{kaplan2019siri} in a workshop held on the campus of Dartmouth College.
At that time, people firmly believe that the efforts of a generation could bring about the realization of artificial intelligence.
However, there was an underestimation of the difficulty of achieving intelligence. 
By 1970, both in academia and the industry, enthusiasm for AI has gradually faded, leading to the emergence of the AI winter.
In nearly 2000, AI once again achieves a breakthrough, accompanied by a series of new highlights, 
including deep blue~\cite{campbell2002deepblue} (the first chess program that beat the human champion built in 1997), Kismet~\cite{Kismet} (can recognize and simulate emotions developed in 2000), Watson~\cite{ferrucci2013watson} (IBM's question answering computer that wins first place on popular 1 million prize television quiz show in 2011), \textit{etc}.

During this time,  coding techniques are experiencing steady growth and evolving towards the direction of complex handcrafted patterns.
Go through the direct entropy coding~\cite{witten1987arithmetic,huffman}, transform coding~\cite{pohlig1980ft,ahmed1974dct}, predictive coding~\cite{wallace1992jpeg}, the hybrid transform/predictive coding routes~\cite{kalva2006h264,fan2004avs} developed as the mainstream of research and standard efforts.
By looking into the signal statistics carefully to develop advanced handcrafted modes, the signal pattern that appears at a large probability is captured to remove the redundancy.

When getting into the deep learning era in 2012,
the models that have strong nonlinear modeling capacities and enable end-to-end training bring in revolutionary changes. 
One typical successful application is in computer vision, achieving significant breakthroughs in visual perception tasks such as classification~\cite{he2016resnet,simonyan2015very}, object detection~\cite{girshick2015fastrcnn,redmon2016yolo}, semantic parsing~\cite{long2015fcn,wang2018ssunderstanding}, depth estimation, and 3D tasks~\cite{laga2022stereodepth,poggi2022bsde}. 
This rapid evolution is closely tied to the development of a wide array of deep learning models,
including
convolutional neural network~\cite{simonyan2015very,poggi2022bsde},
recurrent neural network~\cite{schuster1997brnn,hochreiter1997lstm},
transformer~\cite{liu2021swintrans,han2023vit},
large language model~\cite{devlin2019bert,bao2022vlmo},
diffusion~\cite{dhariwal2021diffusion,yang2023diffusuion},
reinforcement learning~\cite{le2022deeprl,sutton2018rl},
\textit{etc}.

The accuracy of classification and regression tasks in computer vision relies heavily on the predictive/generative capability of models.
As shown in Fig.~\ref{fig:his}, the rapid development of deep learning has propelled a rapid increase in recognition accuracy. 
In 2012, AlexNet~\cite{alex2017alexnet}, representing deep learning methods, surpassed traditional approaches on ImageNet. 
In 2015, ResNet~\cite{he2016resnet} further pushed machine vision perception accuracy beyond that of humans.
On the other hand, model development has also promoted the rapid evolution of coding technologies. 
As shown in Fig.~\ref{fig:his}, over the past 30 years, coding efficiency has doubled approximately every decade,
leading to the emergence of new generations of coding standards. 
However, since the advent of deep learning-based video coding in 2018, after four years of development, 
its performance has surpassed Versatile Video Coding (VVC), with recent work in 2024 surpassing VVC by 30\%.

In recent years, generative model-based applications have rapidly advanced in the field of artificial intelligence.
The influence of generative adversarial networks~\cite{creswell2018ganoverview,goodfellow2020} in the category of the generative model, is rooted in the core idea of game play to address the conditional probability modeling problem.
The process of game play improves the modeling capacities of both the discriminator and generator, with the latter being utilized to produce visually authentic images/videos.
Later on, flow models and diffusion models significantly further improve the quality in image and video generation, \textit{e.g.} stable diffusion~\cite{rombach2022sd} and Dall-E3~\cite{dalle}.
In recent two years, large foundation models have achieved significant success in a series of tasks, including natural language processing (NLP)~\cite{chatgpt}, image/video generation~\cite{midjourney,sora}, image editing~\cite{kawar2023imagic}, \textit{etc.}, with a key property, \textit{i.e.} their capacity to characterize complex distributions.

Recently, the rise of large foundation models also brings new potential to coding techniques.
With the powerful probability modeling capacity, by providing appropriate prompts, these models can sample based on conditional probabilities to generate high-quality results that are visually satisfying and can serve both human visual perception and semantic analysis requirements. 
The transformer architecture is adopted in such large foundation models, which utilize attention mechanisms to learn high-order statistical correlations.
With large-scale datasets, these models can support the estimation of complex posterior probabilities to effectively address the challenge of learning complex long-tail distributions. 
A large amount of parameters aids large foundation models to effectively describe data distributions close to real applicable scenarios.
Through facilitating the probability modeling in high-dimensional space and the generation of images and videos that are consistent with human visual prior and physical priors, the efficiency of coding and reconstruction accuracy are both improved.


\subsection{Multi-Task Modeling}

Over the past two decades, multi-task learning has played an active role all the time in model learning within computer vision and machine learning.
%
%
At an earlier stage, the rise of multi-task learning~\cite{zhang2022mtl} is caused by insufficient training data.
Leveraging multi-task learning enables using more data from diverse tasks, leading to improved performance of the trained model.

\begin{figure}[t]
	\centering
	\includegraphics[width=0.5\textwidth]{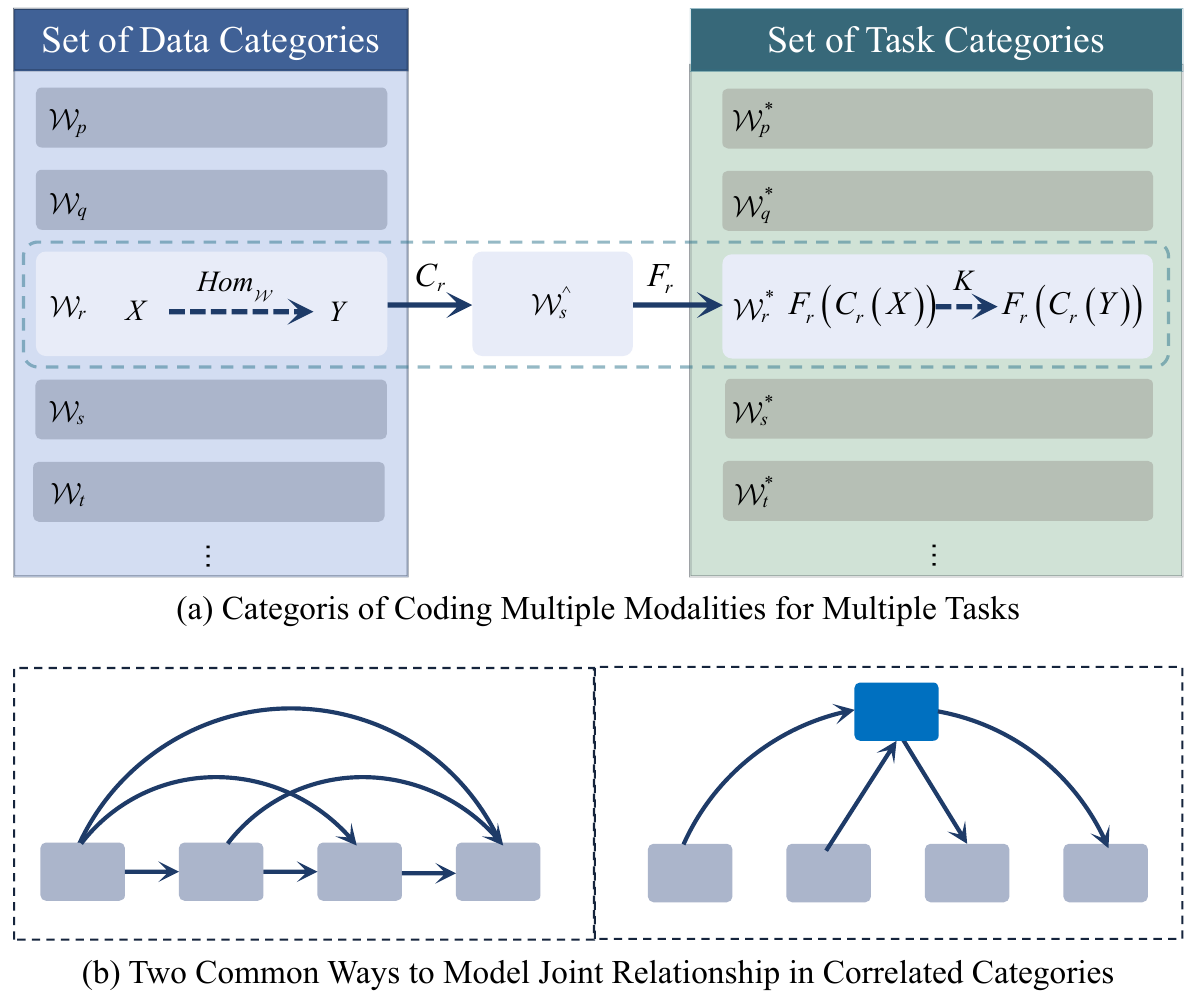}
	\caption{
        (a) The recently emerging CfI instances that extend the categories in the dimensions of data, task, and model
        to form the joint optimization of multiple data/tasks via evolving models.
        (b) The recent technical trends that promote intelligence, 
        which corresponds to the paradigms of modeling $\rm{Hom}$ under the guidance of $K$ in sequential conditional modeling or reparameterization.
	}
	\label{fig:cfi}
\end{figure}

In the era of big data, multi-task learning not only benefits from the abundance of data but also aids in extracting common knowledge by combining multiple tasks.
In detail, multi-task learning can effectively explore more intrinsic feature representation through task association, so that the model can bypass some unreasonable optimization, \wh{alleviate} the risk of overfitting, and \wh{achieve} more intrinsically optimal mapping between the data and label.
\wh{
Due to the development of deep learning, tasks are formulated with the support of flexible composable modules, which could be effectively trained in an end-to-end manner for multi-task optimization.}
During this time, multi-task learning has emerged as a critical role in solving computer vision and natural language processing problems.
Significant improvements have been achieved through modal adaptation~\cite{gabriela2017dava},
utilizing one task to guide the learning of another~\cite{wang2018sft,zhong2023gsr}, or direct multi-task joint optimization~\cite{li2022allinone,vandenhende2022mtldense}.
\wh{
From this point of view, the intrinsic relevance of the multiple tasks is revealed and witnessed, which leads to more powerful models.}

\wh{
The significant impact of multi-task learning is that it induces a paradigm shift in the convergence of model patterns.}
%
\wh{
The backbone networks are built, \textit{e.g.} VGG~\cite{simonyan2015very}, GoogleNet~\cite{szegedy2015googlenet}, ResNet~\cite{he2016resnet}, Transformer~\cite{xiong2020ln}, \textit{etc.}
These models are pre-trained on large-scale datasets and then finetuned on small datasets for various downstream tasks, resulting in improved performance due to the beneficial prior knowledge borrowed from other tasks.
The rise of self-supervised learning marks the critical milestones for the novel paradigm of feature learning that helps handle multiple tasks.
Without the reliance on the ground truth, the model can be trained upon the correspondence of different views of the data itself, which can be trained with massive unlabeled data and offer strong capacities to adapt to a series of downstream tasks.
When the bottlenecks of data, computational power, and task quantity are overcome, large foundation models finally pave their way to versatile task modeling.
By adopting small-scale efficient fine-tuning or prompt techniques, the methods have achieved unprecedented capabilities in few-shot multi-task scenarios.
}
Meanwhile, in the compression domain, coding techniques are also heavily influenced by multi-task learning techniques.
With the development of deep learning that gives birth to both end-to-end training and learnable bitrate estimation, recent efforts turn to the need and route for joint compression of different media for diverse tasks. 
The video coding for machines~\cite{duan2020vcm} raises the question of human-machine collaborative coding, and forms several technical paths, including machine targeted codecs~\cite{Le2021icassp,Choi2020eccv}, feature compression~\cite{Alvar2020icassp, Shah2020icassp}, scalable coding~\cite{wang2021tmm,liu2021scalable}, \textit{etc.}

Despite these works addressing the multi-task modeling issue within the compression scenario, they lack a deeper consideration of the \wh{potential intrinsic synergy and complement/conflict} and how their bitstreams are organized to meet on-demand requests.
The rise of large foundation models brings new chances. 
With the versatile representation as well as the powerful prompt tools provided by the foundation model, coding centered on multiple tasks/media with the assistance of large foundation models can be foreseen and is worth exploring.


\section{Potential Solutions}
\label{sec:trend}
To handle the challenges mentioned above, there are many efforts centering on data, model, and task perspectives.
Based on our discussions derived from the categorical theory,
the challenges of multi-modality data and multi-task modeling are intrinsically equal to modeling and characterization of relationships.
Therefore, recent trends have led to the exploration of two relationship modeling paradigms, similar to the methodology in Section~3.2,
paving the way for new approaches in multi-modality data and multi-task modeling.

As shown in Fig.~\ref{fig:cfi} (b), for recent coding practices,
there are two ways to model $\rm{Hom}$ under the guidance of $K$
in sequential conditional modeling or reparameterization manner.
In the first way, the bitstream of different modalities and tasks is optimized jointly for intelligent bitstream collaboration.
In the second manner, the re-representation of multiple tasks is developed for a more compact and effective bitstream.

%
%
%
%

\subsection{Intelligent Bitstream Collaboration}
%

%
%
%
Considering the source of the matter, as we mentioned in the review part, the world itself, and the related semantics, as well as the objects and scenes, are complex and diverse.
With different forms and characteristics, their features have different distributions and \wh{require various information granularity with diverse bitrates.}
In the context of intelligent analytics, when these different kinds of media/tasks need to be considered jointly, \textit{scalable} naturally becomes a necessary property and the question arises on how to organize and make full use of bitstreams of different methods in a complementary way.
From the data and task perspectives, we can build the connection of multiple tasks and optimize as follows:
\begin{align}
\scriptsize
& \sum\limits_{j < J}^{} {{L_{{\rm{Perf}}}}\left( {{F_j} \left( b_j^i | \theta_{f_j} \right)} \right)}  + \sum\limits_{j < J}^{} {{L_{{\rm{En}}}}\left( b_j^i \right)}   \nonumber \\ \le
& \sum\limits_{ k < J }^{} {{L_{{\rm{Perf}}}}\left( {{F_{ k }} \left(  b_{k}^i \Big| {b_{k-1}^i } , {\theta _{{f_{k}}}} \right) } \right)}  + \sum\limits_{ k < J}^{} {{L_{{\rm{En}}}}\left( b_{k}^i \Big| {b_{k-1}^i }  \right)}, \nonumber 
\end{align}
\begin{align}
\scriptsize
\label{eq:scalable}
& b_{j}^{i} = C_j \left( {x_{j}^{i} \Big | {\theta _{c_j}}} \right), \nonumber \\
& b_{k}^{i} = C_k \left( {x_{k}^{i} \Big | {\theta _{c_{k}}}} \right), \text{\quad} b_{k-1}^{i} = C_{k-1} \left( {x_{k-1}^{i} \Big | {\theta _{c_{k-1}}}} \right),
\end{align}
where $b_{j}^{i}$ defines the compressed bitstream.
This formulation demonstrates an optimization path by linking task sequences, similar to the first case of Fig.~\ref{fig:cfi} (b),
conditioning the current coded bitstream on the previous bitstream, thereby enabling effective joint compression optimization of multiple tasks.
For simplicity, we omit the label in $L_{{\rm{Perf}}}$.

\begin{align}
\label{eq:domain1}
\scriptsize
& \sum\limits_{j < J}^{} {{L_{{\rm{Perf}}}}\left( {{F_j}\left( b_j^i  \right)} \right)}  + \sum\limits_{j < J}^{} {{L_{{\rm{En}}}}\left( b_j^i \right)}   \nonumber \\ \le
& \sum\limits_{{k} < J}^{} {{L_{{\rm{Perf}}}}\left( {F_{{k}}}\left( b_{(k-1) \rightarrow k}^i | \theta _{{f_{{k}}}} \right) \right)} + \sum\limits_{{j^{'}} < J}^{} {{L_{{\rm{En}}}} \left(  b_{(k-1) \rightarrow k}^i  \right)},  \nonumber \\
& b_{(k-1) \rightarrow k}^{i} = C_k \left( {x_{{k}}^i \Big| x_{{k-1}}^i,{\theta _{{c_k}}}} \right).
\end{align}
That is, the redundancy can be first removed at the input level via conditional modeling.

%
%

As shown in Eqn.~\eqref{eq:scalable} and~\eqref{eq:domain1},
the compression results can be achieved via organizing the bitstream conditionally in a sequential manner,
and a solution is derived by optimizing this upper bound.
In the recently emerging VCM efforts~\cite{duan2020vcm,hu2020vcm,liu2021scalable}, scalability plays an important role in \wh{efficiently} integrating bitstreams of multiple tasks in coding tasks.
With the evolving generative models, we can remove inter-task redundancy~\cite{duan2020vcm} by establishing the correlation between features of different tasks, \textit{e.g.} using the feature of one task to predict that of another, forming their joint optimization for multi-task modeling or human-machine collaborative coding.
The work in~\cite{duan2020vcm} provides a general formulation and framework that defines and solves the optimization problem of intelligent bitstream organization through scalability with two specific exemplar frameworks.
%
%

\begin{table*}[]
	\footnotesize
	\caption{Performance comparisons from both the perspectives of human and machine visions of different scalable coding paradigms.
	}
	\label{tab:scalable}
	\begin{tabular}{c|c|c|c|cc|cc}
		\hline
		Research                     & Bridged Object                                                                         & Method              & -                            & \multicolumn{2}{c}{Human Vision}     & \multicolumn{2}{|c}{Machine Vision} \\
		\hline
		\multirow{5}{*}{Yang \textit{et al.}~\cite{yang2021tmm}} & \multirow{5}{*}{\begin{tabular}[c]{@{}c@{}}Handcrafted Features,\\ Image\end{tabular}} & -                   & Bitrate & LPIPS & SSIM    & NME               & Memorability   \\ \cline{3-8}
		&                                                                                        & Yang et al. (N=15)  & 0.152                        & 0.208                      & 0.706   & 3.042             & 0.900          \\
		&                                                                                        & WEBP (qp=0.4)       & 0.154                        & 0.215                      & 0.798   & 4.392             & 0.880          \\
		&                                                                                        & Yang et al. (N=122) & 0.191                        & 0.165                      & 0.759   & 2.748             & 0.898          \\
		&                                                                                        & WEBP (qp=3)         & 0.197                        & 0.174                      & 0.827   & 3.430             & 0.886          \\
		\hline
		\multirow{7}{*}{Choi \textit{et al.}~\cite{choi2022scalable}} & \multirow{7}{*}{Deep Intermediate Features}                                            & -                   & -                            & BD-PSNR                    & BD-SSIM & BD-mAP (Detect.)  & BD-mAP (Seg.)  \\
        \cline{3-8}
		  &   & VVC                   & -   & 10.17\%     & -7.83\%   & 2.79    & 2.33  \\
            &   & HEVC                  & -   & -14.27\%    & --26.15\% & 4.55    & 3.05  \\
            &   & \cite{minnen2018ar}   & -   & -3.58\%     & -7.83\%   & 3.26    & 3.73  \\
            &   & \cite{Cheng2020}      & -   &  4.49\%     & -1.90\%   & 2.89    & 3.62  \\
            &   & \cite{hu2020coarse}   & -   & -5.58\%     & -16.84\%  & 1.99    & 2.80  \\
            &   & \cite{lee2019context} & -   & -7.01\%     & -15.23\%  & 2.93    & 3.66  \\
		\hline
		\multirow{5}{*}{Yang \textit{et al.}~\cite{yang2023manifold}} & \multirow{5}{*}{Image Manifold, Image}                                                 & -                   & Bitrate                      & LPIPS$_\text{Alex}$                  & LPIPS$_\text{SQ}$ & Mean IU (Seg.)    & Acc. (Class.)  \\
		\cline{3-8}
		&                                                                                        & Yang et al.         & 0.0222                       & 0.1302                     & 0.1172  & 0.7421            & 98             \\
		&                                                                                        & Yang et al.         & 0.0285                       & 0.1287                     & 0.1116  & 0.7422            & 97             \\
		&                                                                                        & VTM (QP=40)         & 0.0504                       & 0.1991                     & 0.1955  & 0.7104            & 96             \\
		&                                                                                        & VTM (QP=42)         & 0.0378                       & 0.2272                     & 0.2193  & 0.7150            & 95             \\
		\hline    
	\end{tabular}
\end{table*}

A performance summary of three typical methods is summarized in Table~\ref{tab:scalable}.
Yang \textit{et al.}~\cite{yang2021tmm} build the scalable mapping from handcrafted features and images.
The evaluations are conducted on VGGFace2~\cite{cao2018vggface2}, including 39,122 images for
training and 3,000 images for evaluation.
Choi \textit{et al.}~\cite{choi2022scalable} connect different layers of deep intermediate features.
In the experimental evaluation, the training data includes CLIC~\cite{clic} and JPEG-AI~\cite{jpegai}, and the testing data includes COCO2014, COCO2017~\cite{lin2014coco}, and Kodak~\cite{kodak}.
Yang \textit{et al.}~\cite{yang2023manifold} create the connection between the pretrained natural image manifold and the image.
The experiments are conducted on FFHQ-Aging~\cite{orel2020lifespan}.
It is observed from Table~\ref{tab:scalable} that, except for the conventional fidelity measures, \textit{e.g.} PSNR and SSIM, scalable coding methods outperform competitive conventional codecs in both perceptual driven measures, \textit{e.g.} LPIPS, and machine vision performance, \textit{i.e.} the performance of downstream tasks.
%
Despite making significant progress, there are remaining issues. First, theoretically, in-depth research is still very limited.
%
In addition, in scalable coding, splitting the bitstream leads to a narrowed context, which inevitably affects the coding efficiency.
It is still an open question about how to obtain a more flexible and separable bitstream while maintaining coding efficiency.

Recently, the rise of large foundation models has brought new changes to multi-modality data compression.
They are capable of modeling data from diverse sources and different modalities uniformly. 
They map those diverse data into a unified feature space.
Dynamic and efficient utilization of multi-modality information through fine-tuning~\cite{hu2022lora,ding2023peft} or prompt learning~\cite{Khattak_2023_ICCV,li2023common} combined with pretrained large foundation models.
Thus, it has great potential to further explore multi-modality data compression theories and approaches shown in Eqn.~\eqref{eq:scalable} with the isomorphism constraint on the bitstreams of different modalities that makes the space of $b_j^i$ for different $j$ containing homomorphically and convertible information.
Namely, these features are mapped into the same space and can be transformed nearly without information loss between each other.


\subsection{Compressive Analytics}

With the widespread construction of collaborative infrastructure and the rise of large foundation models,
multi-task modeling has become a trend.
In the field of compression, recent work has focused on compression of multiple tasks.
Instead of coding images for downstream tasks, also different from intelligent bitstream collaboration,
recent solutions aim to seek general alternative compact features, similar to the idea of intrinsic semantics to describe the relationship of objects,
Formally, the rough idea is illustrated in Eqn~\eqref{eq:multi}:
\begin{align}
\label{eq:multi}
\scriptsize
& \sum\limits_{j < J}^{} {{L_{{\rm{Perf}}}}\left( {{F_j}\left( b_j^i  \right)} \right)}  + \sum\limits_{j < J}^{} {{L_{{\rm{En}}}}\left( b_j^i \right)}   \nonumber \\ \le
& \sum\limits_{{k} < J}^{} {{L_{{\rm{Perf}}}}\left( {F_{{k}}}\left( b_{J \rightarrow k}^i | \theta _{{f_{{k}}}} \right) \right)} + \sum\limits_{{k} < J}^{} {{L_{{\rm{En}}}} \left(  b_{J \rightarrow k}^i  \right)},  \nonumber \\
& b_{J \rightarrow k}^i = C_k \left( {x_{{k}}^i \Big| x_J^i,{\theta _{c_k}}} \right).
\end{align}
A common compact representation is extracted or several representations are organized in a scalable way.

\begin{table*}[htbp]
	\footnotesize
	\centering
	\caption{Performance comparisons in the joint bit-rate of different schemes to cluster multiple tasks. \\
	The best results are denoted in bold.
        SC: Scene Class.
        SS: Semantic Segmentation.
        OC: Object Class.
        N: Normal.
        R: Reshading.
        C: Curvature.
	}
        \resizebox{\textwidth}{!}{
	\begin{tabular}{c|c|cc|cccc|ccccccc}
        \hline
             -          &   -       &   \multicolumn{2} {c|} {Feature Compression}  &   \multicolumn{4} {c|} {Image Compression}     &   \multicolumn{4} {c} {Compressive Analytics}    \\
	\hline
	Task                         & Raw            & Hyperprior~\cite{chamain2021end} & Cheng2020~\cite{Cheng2020} & Le2021-2~\cite{le2021ICM} & BPG-51 & BPG-47 & 
        BPG-43 & Customized & Trinity~\cite{yang2024compresstaskonomy} & {Trinity*}~\cite{yang2024compresstaskonomy} & Hex \\ 
        \hline
	SC (Acc.$\uparrow$)          & 67.48\%     & 48.73\%  & 55.08\% &  62.30\%  & 48.73\% & 56.64\% & 61.91\% & 52.25\% & 53.12\%  &  46.39\%    & 50.98\%   \\
		
	SS (mIoU$\uparrow$)          & 27.65\%     & 30.94\%  & 34.36\% &  26.72\%  & 19.84\% & 24.36\% & 25.37\% & 25.71\% & 28.29\%  &  30.04\%    & 26.70\%   \\
		
	OC (Acc.$\uparrow$)          & 53.61\%     & 43.26\%  & 48.54\% &  51.46\%  & 46.39\% & 50.10\% & 51.07\% & 49.90\% & 48.93\%  &  45.31\%    &  42.77\%  \\
		
	N  ($L_1$ $\downarrow$)      & 0.121       & 0.136    & 0.134   &  0.132    & 0.229   & 0.176 & 0.147     & 0.129   & 0.135    &  0.133      &  0.133   \\
		
	R  ($L_1$ $\downarrow$)      & 0.283       & 0.244    & 0.241   &  0.254    & 0.410   & 0.334 & 0.305     & 0.231   & 0.245    &  0.248      &  0.225   \\
		
    C  ($L_1$ $\downarrow$)      & 0.384       & 0.391    & 0.392   &  0.397    & 0.418   & 0.402 & 0.394     & 0.389   & 0.391    &  0.387      &  0.392   \\
	\hline
	Bit-Rate (Bpp $\downarrow$)  & -           & 0.049    & 0.188   &   0.334   & 0.024   & 0.039 & 0.063     & 0.078   & 0.042    &  0.053      &  0.059   \\
	\hline
	\end{tabular}
        }
	\label{tab:mt}%
	\vspace{-2mm}
\end{table*}

%
%

%
%

%
%


The work of compressive taskonomy~\cite{yang2024compresstaskonomy} makes a comprehensive study of existing compression paradigms for analytics,
including 
machine vision targeted codec~\cite{le2021ICM,huang2021RDICM},
feature compression~\cite{chen2020intermediate,chen2019lossy,alvar2021bitallo,zhang2021msfc},
scalable coding~\cite{liu2021scalable,Prabhakar2021dcc}, and
compressive analytics~\cite{yang2024compresstaskonomy}.
The performance comparisons of those typical methods are summarized in Table~\ref{tab:mt}.
The experiments are conducted on the Taskonomy dataset~\cite{zamir2018taskonomy}.
A subset is selected at random, including 51,316 images for training, 945 for validation, and 1,024 for testing.
The performance for semantic segmentation is reported with the measures of mean pixel-level accuracy~(Acc.), the accuracy of pixels in the non-background regions~(Non-BG Acc.), and mean IoU~(mIoU), Cross Entropy Loss.
The exploration leads to a few interesting insights: 
\textit{1) feature compression methods, especially the learned-based ones (Hyperprior~\cite{chamain2021end} and Cheng2020~\cite{Cheng2020} for feature compression), are more powerful than codecs (Le2021-1~\cite{Le2021icassp} and Le2021-2~\cite{le2021ICM}) that compress images for intelligent analytics;}
\textit{2) we can observe the space for further improvement of the existing scalable coding route and the existing usual setting involving the visual reconstruction have a conflict with and might mislead the performance of analytics.}

This work also studies what strategies will be more effective when facing the scenario of compressing multiple tasks or multi-modality input.
The results are shown in Table~\ref{tab:mt}.
Several methods are compared:
\begin{itemize}
	\item Customized: compressing each task independently;
	\item Hex: jointly compressing all representations together with one model;
	\item Trinity: compressing two ideal groups;
	\item Trinity*: compressing two groups different from those of Trinity.
\end{itemize}

Two interesting insights are reached.
\textit{First, it is demonstrated the existence of the dimension/information gap affects the performance and the incorporation of specifically designed task-relevant modules, \textit{e.g.} codebook hyperprior~\cite{yang2024compresstaskonomy}, can bridge the gap and can naturally lead to improved performance (Hyperprior~\cite{chamain2021end} vs. Trinity~\cite{yang2024compresstaskonomy}).}
\textit{Second, it is observed that, different ways of task clustering lead to different performances, and clustering intrinsically similar tasks as a group can lead to better compression performance (Customized vs. Trinity vs. Trinity* vs. Hex).}

All these results inspire the community to further study the task relationship in depth in future work,
including the intrinsic task similarity, correlation of their feature distribution, \textit{etc.}, to facilitate designing more effective and generalizable models for intelligent analytics.

\section{Discussions}
\label{sec:diss}
Despite considerable efforts, there are still many challenges and opportunities on the path, 
with coding being an important pathway.
This section discusses the significance of large foundation models from the coding perspective, 
while also revealing that the gap between $K$ and $L$ may be a potential obstacle along the way.

\begin{figure}[t]
	\centering
	\includegraphics[width=0.5\textwidth]{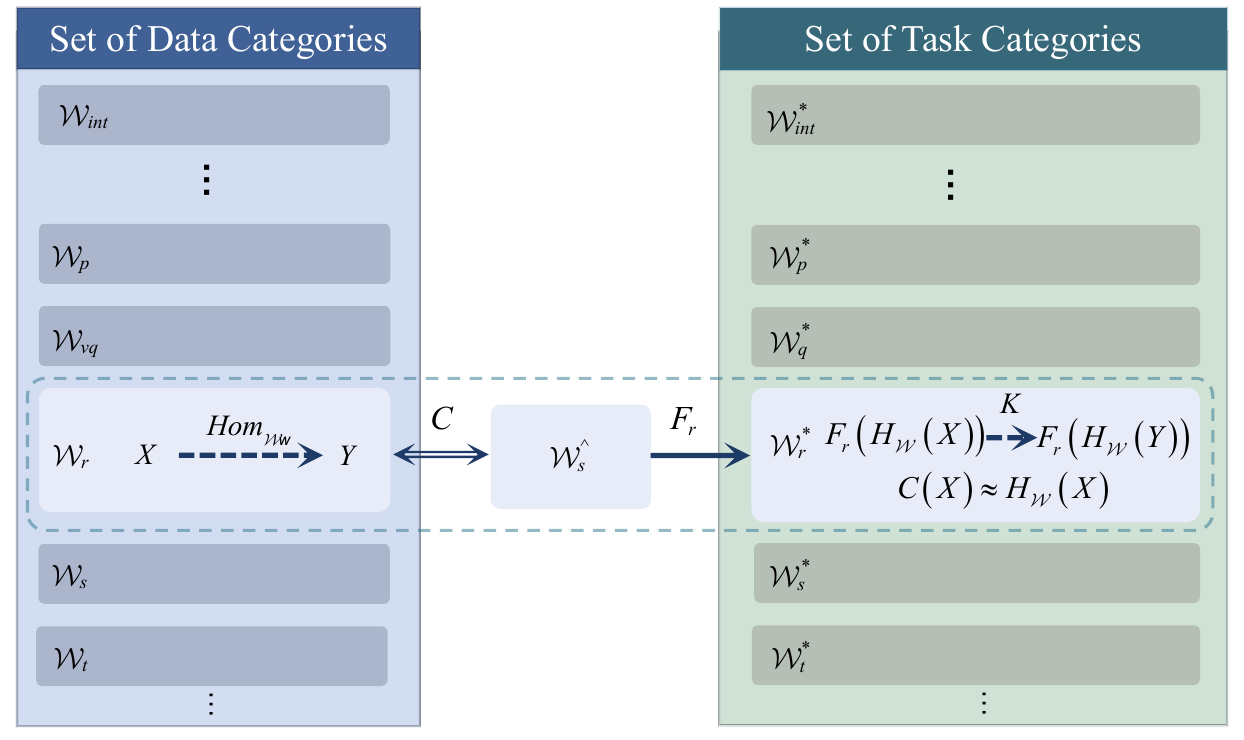}
	\caption{
        The potential explanation for scaling law of large foundation model in category theory.
        In large foundation models, $\mathcal{W}_\text{int}$ and $\mathcal{W}_\text{int}^{*}$, categories that can approximately capture intrinsic semantic relationships, can be derived with the power of huge-scale models.
        In this case, $C(X)$ approximates extracting intrinsic semantics $H_{\mathcal{W}}$ ideally,
        and downstream tasks equal to acquiring the sub-relationship from the $C(X)$, which leads to improved generalization and robustness.
	}
	\label{fig:dis}
\end{figure}

\subsection{Large Foundation Model: Necessary Route for Diverse Tasks/Domains}
The rise of large foundation models has indeed led to a practical improvement in model performance.
However, the accompanying massive computational requirements have also raised concerns.
Our theoretic framework can explain that.
For convenience, we regard the downstream tasks as the same and ignore them in the formulation.
The benefit and cost of using large foundation models or small specified models can be represented as follows:

\begin{scriptsize}
\begin{align}
U_{s} = \displaystyle\lim_{J \rightarrow \infty} \sum\limits_{j < J}^{} \left\{ \sum\limits_{0 < i \le {N_j}}^{} \left( {{L_{{\rm{P}}}}\left( {{F_j}\left( {
b_j^i |{\theta _{{f_j}}}} \right) } \right)}  + {{L_{{\rm{E}}}}\left( b_j^i \right)} \right) + {{L_{{\rm{C}}}}\left( {{\theta _{{c_j}}}} \right)} \right\}, \nonumber
\end{align}
\end{scriptsize}

\begin{scriptsize}
\begin{align}
\label{eq:ua}
& U_{a}  = \displaystyle\lim_{J \rightarrow \infty} \sum\limits_{j < J}^{} \left\{ \sum\limits_{0 < i \le {N_j}}^{} \left( {{L_{\rm{P}}}\left( {{F_j}\left( {
b_{j|a}^i |{\theta _{{f_j}}}} \right) } \right)}  + {{L_{\rm{E}}}\left( b_{j|a}^i \right)} \right) \right\} + {{L_{\rm{C}}}\left( {{\theta _{{c_a}}}} \right)}, \nonumber \\
& b_{j|a}^{i}  = C_a \left( {x_{j}^{i} \Big | {\theta _{c_a}}} \right),
\end{align}
\end{scriptsize}

\noindent where $L_{\rm{P}}$, $L_{\rm{E}}$, and $L_{\rm{C}}$ short for  $L_{\rm{Perf}}$, $L_{\rm{En}}$, and $L_{\rm{Com}}$, respectively.
$C_a$ denotes a universal coding module, which might be close to ideal coding provided by large foundation models, 
while $b_{j | a}^{i}$ is the bitstream formed by $C_a$.
When $J \rightarrow \infty$, the comparison between $U_{s}$ and $U_{a}$ are dominated by the comparisons of ${{L_{{\rm{Com}}}}\left( {{\theta _{{c_a}}}} \right)}$ and $\sum\limits_{j < J}^{} {{L_{{\rm{Com}}}}\left( {{\theta _{{c_j}}}} \right)}$.
In this case, using a large-scale $\theta _{{c_a}}$ leads to an overall lower cost.
That is to say, faced with an infinite number of tasks/applications, 
aggregating resources to train large models and then adapting them to specific tasks is far superior to developing small models for each scenario.

This viewpoint is also verified in a recent work~\cite{huh2024platonic}. 
Deep networks tend to seek simple fitting regularity to data, and this inherent simplicity bias makes models simplify their representations, 
\textit{i.e.} coding data to form knowledge.
Besides, models with different architectures and goals tend to converge on a consistent underlying representation.
\wh{
This observation is an embodiment of the Yoneda Lemma: since data of different modalities/views are produced by the same objects,
their relationships $K$ in different feature spaces actually aligned.
It also reveals to extent possibility of the existence of optimal coding ways,
which coincides with our Ideal Coding definition, with the potential implementation way of large foundation models.
}

\subsection{Connection and Dilemma of Modeling Objective}
Our framework unifies the exploration of several fields from the category theory perspective.
Their difference mainly lies in $K$. $K$ actually defines the task and their relationship. 
The recent works demonstrate that, some commonly used $L$ are homologous.
For example, in~\cite{rosasco2004same}, the potential relationship between the hinge loss, logistic loss, and square loss is discussed.
In~\cite{zhou2022allloss}, from the Neural Collapse perspective,  losses including cross-entropy loss, label smoothing, focal loss, and MSE are demonstrated to produce equivalent features on training data.
In~\cite{nowozin2016fgan}, generative adversarial learning can be categorized into general variational divergence
estimation approach and a series of f-divergence can be used for training generative neural samplers.
Some works also demonstrate the potential conflict~\cite{blau2018pdt,cohen2024the} between different measures for low-level vision tasks, \textit{e.g.} perceptual driven measures, \textit{e.g.} Peak-Signal-to-Noise Ratio (PSNR), and Learned Perceptual Image Patch Similarity (LPIPS).
Whether there is a way to effectively combine the advantages of these two kinds or make a good trade-off between them is still an open question.
From a fundamental methodological perspective, using $L$ to describe $K$ is a commonly used empirical attempt. 
Whether there is a theoretical guarantee, and whether there exists $K$ that cannot be described in concrete mathematical forms, are also unexplored questions awaiting further investigation.

Additionally, in the paradigm of large foundation models,
the method of modeling $Hom$ under the guidance of $K$ mostly adopts the form shown in Fig.~\ref{fig:cfi} (a), \textit{e.g.} GPT~\cite{brown2020gpt} and LLAMA~\cite{touvron2023llama}.
These methods have achieved great success through unsupervised autoregressive learning based on a vast number of parameters.
However, such sequential prediction structures of this architecture also limit the model's context size, leading to inference limits due to the constrained local information.
The recent attempt~\cite{tian2024var}, combining the core ideas of both Fig.~\ref{fig:cfi} (a) and Fig.~\ref{fig:cfi} (b), replaces the next token prediction with the next scale prediction.
Therefore, when predicting the next scale information, the model can acquire the related information of hierarchical scales, which adapts the critical properties of LLMs, \textit{i.e.} scaling laws and generalization, into the vision modality.

%
%

%
%

%
%

\subsection{Emergent Capacities: In the Way to Ideal Coding}
%
Currently, an interesting and compelling issue is the explanation of scaling laws.
We found that, that open issue might well align with the way towards ideal coding.
%
As shown in Fig.~\ref{fig:dis}, the collected data is the sample of the intrinsic scene
and the task represents a view of intrinsic semantics.
An intuitive explanation is that, when the model's capacity is lower than a necessary bound, 
it will fail to reach the category of intrinsic scene or intrinsic semantic.
The model just relies on statistical modes to fit the data correlation among different tasks.
Once the models own a certain capacity, being able to reach the intrinsic scene and intrinsic semantics,
the models can somehow learn the real intrinsic mapping relationship, thus having significantly improved generalization and robustness.

Another potential explanation is the large foundation models, with the help of diverse data/tasks, can be regarded as a combination of meta-learning and model learning.
The task relationship as well as their organization are learned implicitly in a joint manner 
and the obtained model naturally learns to be generalizable to handle diverse kinds of tasks.
The work in~\cite{huh2024platonic} also demonstrates that, as the model size and performance increase, 
the alignment of representations across different modalities and tasks becomes more apparent.
Therefore, our work and such platonic representation provide attempts at both theoretical and practical levels in a complementary way.

\subsection{Summary}
Thus far, we have presented the theoretical foundation of modeling coding problems to achieve intelligence in a general way from the perspective of category theory.
We also have utilized MDL to transform the theory into a computationally solvable problem, opening up new avenues
for the future integration of coding and intelligence technologies.
At the same time, recent studies also corroborated our theoretically derived conjectures and results from technical practice perspectives.
OpenAI demonstrates the equivalence of compression and model learning via the practice of large foundation models.
The platonic representation~\cite{huh2024platonic} shows the convergence of different model architecture and multi-modality leads to the same intrinsic semantic representation.
Therefore, we found that in various views, surprisingly these works complement and corroborate each other, leading to the same destination. 

%
%







\section{Conclusion}
\label{sec:con}

In today's landscape of deep integration between coding technology and artificial intelligence,
this paper formulates a novel problem of \textit{Coding for Intelligence} from the category theory view, providing a general framework to unify coding and model learning.
Under this unified understanding framework, key elements and critical routes are identified with potential solutions, preliminary results, and the future direction of CfI presented.
The state-of-the-art techniques centering on multi-task/modality collaboration from the perspective of data, object, and task, are reviewed comprehensively. 
As an initial attempt to identify the roles and principles of CfI, this work is expected to call for more
efforts from both academia and industry, to facilitate the fusion of coding and AI and widely utilization of AI power.

\footnotesize
\bibliographystyle{IEEEtran}
\bibliography{cfi}

\begin{thebibliography}{100}
\providecommand{\url}[1]{#1}
\csname url@samestyle\endcsname
\providecommand{\newblock}{\relax}
\providecommand{\bibinfo}[2]{#2}
\providecommand{\BIBentrySTDinterwordspacing}{\spaceskip=0pt\relax}
\providecommand{\BIBentryALTinterwordstretchfactor}{4}
\providecommand{\BIBentryALTinterwordspacing}{\spaceskip=\fontdimen2\font plus
\BIBentryALTinterwordstretchfactor\fontdimen3\font minus \fontdimen4\font\relax}
\providecommand{\BIBforeignlanguage}[2]{{%
\expandafter\ifx\csname l@#1\endcsname\relax
\typeout{** WARNING: IEEEtran.bst: No hyphenation pattern has been}%
\typeout{** loaded for the language `#1'. Using the pattern for}%
\typeout{** the default language instead.}%
\else
\language=\csname l@#1\endcsname
\fi
#2}}
\providecommand{\BIBdecl}{\relax}
\BIBdecl

\bibitem{sternberg2011cognitive}
R.~Sternberg, K.~Sternberg, and J.~Mio, \emph{Cognitive Psychology}.\hskip 1em plus 0.5em minus 0.4em\relax Wadsworth/Cengage Learning, 2011.

\bibitem{tyler2010psy}
A.~Tyler, \emph{Psychological Types}.\hskip 1em plus 0.5em minus 0.4em\relax Boston, MA: Springer US, 2010, pp. 723--725.

\bibitem{dedre1983analogy}
D.~Dedre, ``Structure-mapping: A theoretical framework for analogy,'' \emph{Cognitive Science}, vol.~7, no.~2, pp. 155--170, 1983.

\bibitem{banich2010generalization}
M.~T. Banich and D.~Caccamise, Eds., \emph{Generalization of Knowledge: Multidisciplinary Perspectives}, 1st~ed.\hskip 1em plus 0.5em minus 0.4em\relax Psychology Press, 2010.

\bibitem{brady2008memory}
T.~F. Brady, T.~Konkle, G.~A. Alvarez, and A.~Oliva, ``Visual long-term memory has a massive storage capacity for object details,'' \emph{Proceedings of the National Academy of Sciences}, vol. 105, no.~38, pp. 14\,325--14\,329, 2008.

\bibitem{birbaumer1999spelling}
N.~Birbaumer, N.~Ghanayim, T.~Hinterberger, I.~Iversen, B.~Kotchoubey, A.~Kübler, J.~Perelmouter, E.~Taub, and H.~Flor, ``A spelling device for the paralysed,'' \emph{Nature}, vol. 398, no. 6725, pp. 297--298, Mar 1999.

\bibitem{reed2000enc}
S.~K. Reed, ``Thinking: Problem solving,'' in \emph{Encyclopedia of psychology}, A.~E. Kazdin, Ed.\hskip 1em plus 0.5em minus 0.4em\relax Oxford University Press, 2000, vol.~8, pp. 71--75.

\bibitem{johnson2000reason}
P.~N. Johnson-Laird, ``Thinking: Reasoning,'' in \emph{Encyclopedia of psychology}, A.~E. Kazdin, Ed.\hskip 1em plus 0.5em minus 0.4em\relax Oxford University Press, 2000, vol.~8, pp. 75--79.

\bibitem{russell2016artificial}
S.~J. Russell and P.~Norvig, \emph{Artificial Intelligence: a modern approach}, 3rd~ed.\hskip 1em plus 0.5em minus 0.4em\relax Pearson, 2009.

\bibitem{dayan2005neuroscience}
P.~Dayan and L.~F. Abbott, \emph{Theoretical Neuroscience: Computational and Mathematical Modeling of Neural Systems}.\hskip 1em plus 0.5em minus 0.4em\relax The MIT Press, 2005.

\bibitem{squire2012neuroscience}
L.~Squire, D.~Berg, F.~Bloom, S.~{Du Lac}, A.~Ghosh, and N.~Spitzer, \emph{\BIBforeignlanguage{English (US)}{Fundamental Neuroscience: Fourth Edition}}.\hskip 1em plus 0.5em minus 0.4em\relax Elsevier Inc., November 2012.

\bibitem{roozendaal2002neurobiology}
B.~Roozendaal, ``Stress and memory: Opposing effects of glucocorticoids on memory consolidation and memory retrieval,'' \emph{Neurobiology of Learning and Memory}, vol.~78, no.~3, pp. 578--595, 2002.

\bibitem{roozendaal2003neuro}
------, ``Systems mediating acute glucocorticoid effects on memory consolidation and retrieval,'' \emph{Progress in Neuro-Psychopharmacology and Biological Psychiatry}, vol.~27, no.~8, pp. 1213--1223, 2003.

\bibitem{metcalfe1986feelingok}
J.~Metcalfe, ``Feeling of knowing in memory and problem solving,'' \emph{Journal of Experimental Psychology: Learning, Memory and Cognition}, vol.~12, pp. 288--294, 1986.

\bibitem{srivastava2014drouput}
N.~Srivastava, G.~Hinton, A.~Krizhevsky, I.~Sutskever, and R.~Salakhutdinov, ``Dropout: A simple way to prevent neural networks from overfitting,'' \emph{Journal of Machine Learning Research}, vol.~15, no.~56, pp. 1929--1958, 2014.

\bibitem{doersch2016vae}
C.~{Doersch}, ``{Tutorial on Variational Autoencoders},'' \emph{arXiv e-prints}, p. arXiv:1606.05908, June 2016.

\bibitem{ma2015avs2}
S.~Ma, T.~Huang, C.~Reader, and W.~Gao, ``{AVS2? Making Video Coding Smarter [Standards in a Nutshell]},'' \emph{{IEEE} Signal Processing Magazine}, vol.~32, no.~2, pp. 172--183, 2015.

\bibitem{bross2021vvc}
B.~Bross, Y.-K. Wang, Y.~Ye, S.~Liu, J.~Chen, G.~J. Sullivan, and J.-R. Ohm, ``Overview of the versatile video coding {(VVC)} standard and its applications,'' \emph{{IEEE} Trans. on Circuits and Systems for Video Technology}, vol.~31, no.~10, pp. 3736--3764, 2021.

\bibitem{simonyan2015very}
K.~Simonyan and A.~Zisserman, ``Very deep convolutional networks for large-scale image recognition,'' in \emph{Proc.~Int'l Conf.~Learning Representations}, 2015, pp. 1--14.

\bibitem{he2016resnet}
K.~He, X.~Zhang, S.~Ren, and J.~Sun, ``Deep residual learning for image recognition,'' \emph{Proc.~IEEE Int'l Conf.~Computer Vision and Pattern Recognition}, pp. 770--778, 2016.

\bibitem{girshick2015fastrcnn}
R.~Girshick, ``Fast r-cnn,'' in \emph{Proc.~IEEE Int'l Conf.~Computer Vision}, 2015, pp. 1440--1448.

\bibitem{redmon2016yolo}
J.~Redmon, S.~Divvala, R.~Girshick, and A.~Farhadi, ``You only look once: Unified, real-time object detection,'' in \emph{Proc.~IEEE Int'l Conf.~Computer Vision and Pattern Recognition}, 2016, pp. 779--788.

\bibitem{long2015fcn}
J.~Long, E.~Shelhamer, and T.~Darrell, ``Fully convolutional networks for semantic segmentation,'' in \emph{Proc.~IEEE Int'l Conf.~Computer Vision and Pattern Recognition}, 2015, pp. 3431--3440.

\bibitem{noh2017retrieval}
H.~Noh, A.~Araujo, J.~Sim, T.~Weyand, and B.~Han, ``Large-scale image retrieval with attentive deep local features,'' in \emph{Proc.~IEEE Int'l Conf.~Computer Vision}, 2017, pp. 3476--3485.

\bibitem{touvron2023llama}
H.~Touvron, L.~Martin, K.~Stone \emph{et~al.}, ``Llama 2: Open foundation and fine-tuned chat models,'' \emph{arXiv preprint arXiv:2307.09288}, 2023.

\bibitem{kirillov2023segany}
A.~Kirillov, E.~Mintun, N.~Ravi, H.~Mao, C.~Rolland, L.~Gustafson, T.~Xiao, S.~Whitehead, A.~C. Berg, W.-Y. Lo, P.~Doll{\'a}r, and R.~Girshick, ``Segment anything,'' \emph{arXiv:2304.02643}, 2023.

\bibitem{evaclip}
Q.~Sun, Y.~Fang, L.~Wu, X.~Wang, and Y.~Cao, ``Eva-clip: Improved training techniques for clip at scale,'' \emph{arXiv preprint arXiv:2303.15389}, 2023.

\bibitem{huang2023densityprompt}
Q.~Huang, X.~Dong, D.~Chen, W.~Zhang, F.~Wang, G.~Hua, and N.~Yu, ``Diversity-aware meta visual prompting,'' in \emph{Proc.~IEEE Int'l Conf.~Computer Vision and Pattern Recognition}, June 2023, pp. 10\,878--10\,887.

\bibitem{liu2023explicit}
W.~Liu, X.~Shen, C.-M. Pun, and X.~Cun, ``Explicit visual prompting for low-level structure segmentations,'' in \emph{Proc.~IEEE Int'l Conf.~Computer Vision and Pattern Recognition}, 2023, pp. 19\,434--19\,445.

\bibitem{deletang2023language}
G.~Delétang, A.~Ruoss, P.-A. Duquenne, E.~Catt, T.~Genewein, C.~Mattern, J.~Grau-Moya, L.~K. Wenliang, M.~Aitchison, L.~Orseau, M.~Hutter, and J.~Veness, ``Language modeling is compression,'' \emph{arXiv preprint arXiv:2309.10668}, 2023.

\bibitem{duan2020vcm}
L.-Y. Duan, J.~Liu, W.~Yang, T.~Huang, and W.~Gao, ``Video coding for machines: A paradigm of collaborative compression and intelligent analytics,'' \emph{{IEEE} Trans. on Image Processing}, vol.~29, pp. 8680--8695, 2020.

\bibitem{hu2020vcm}
Y.~Hu, S.~Yang, W.~Yang, L.-Y. Duan, and J.~Liu, ``Towards coding for human and machine vision: A scalable image coding approach,'' in \emph{Proc.~IEEE Int'l Conf.~Multimedia and Expo}, 2020, pp. 1--6.

\bibitem{yang2024compresstaskonomy}
W.~Yang, H.~Huang, Y.~Hu, L.-Y. Duan, and J.~Liu, ``Video coding for machines: Compact visual representation compression for intelligent collaborative analytics,'' \emph{{IEEE} Trans. on Pattern Analysis and Machine Intelligence}, 2024.

\bibitem{legg2007ui}
S.~Legg and M.~Hutter, ``Universal intelligence: A definition of machine intelligence,'' \emph{Minds and Machines}, vol.~17, no.~4, pp. 391--444, 12 2007.

\bibitem{bickley1995threestratum}
P.~G. Bickley, T.~Z. Keith, and L.~M. Wolfle, ``The three-stratum theory of cognitive abilities: Test of the structure of intelligence across the life span,'' \emph{Intelligence}, vol.~20, no.~3, pp. 309--328, 1995.

\bibitem{davis2011mi}
\BIBentryALTinterwordspacing
K.~Davis, J.~Christodoulou, S.~Seider, and H.~Gardner, ``The theory of multiple intelligences,'' in \emph{The Cambridge Handbook of Intelligence}, R.~J. Sternberg and S.~B. Kaufman, Eds.\hskip 1em plus 0.5em minus 0.4em\relax Cambridge University Press, 2011, pp. 485--503. [Online]. Available: \url{https://doi.org/10.1017/CBO9780511977244.025}
\BIBentrySTDinterwordspacing

\bibitem{sternberg1985biq}
R.~J. Sternberg, \emph{Beyond IQ: A Triarchic Theory of Human Intelligence}.\hskip 1em plus 0.5em minus 0.4em\relax Cambridge University Press, 1985.

\bibitem{hofstadter2001analogy}
D.~Hofstadter, ``Epilogue: Analogy as the core of cognition,'' in \emph{The Analogical Mind: Perspectives from Cognitive Science}.\hskip 1em plus 0.5em minus 0.4em\relax MIT Press, 2001, pp. 499--538.

\bibitem{Wiki2024abs}
``Abstraction --- {W}ikipedia{,} the free encyclopedia,'' \url{https://en.wikipedia.org/wiki/Abstraction}, Accessed 2024-4-17.

\bibitem{gluck2007learning}
\BIBentryALTinterwordspacing
M.~Gluck, E.~Mercado, and C.~Myers, \emph{Learning and Memory: From Brain to Behavior}.\hskip 1em plus 0.5em minus 0.4em\relax Worth Publishers, 2007. [Online]. Available: \url{https://books.google.com/books?id=_VwkAAAAQBAJ}
\BIBentrySTDinterwordspacing

\bibitem{clark1977psychology}
\BIBentryALTinterwordspacing
H.~Clark and E.~Clark, \emph{Psychology and Language: An Introduction to Psycholinguistics}.\hskip 1em plus 0.5em minus 0.4em\relax Harcourt Brace Jovanovich, 1977. [Online]. Available: \url{https://books.google.com/books?id=XlArQAAACAAJ}
\BIBentrySTDinterwordspacing

\bibitem{medin1981}
D.~L. Medin and E.~E. Smith, ``Strategies and classification learning,'' \emph{Journal of Experimental Psychology: Human Learning and Memory}, vol.~7, no.~4, pp. 241--253, 1981.

\bibitem{collins1969}
A.~M. Collins and M.~R. Quillian, ``Retrieval time from semantic memory,'' \emph{Journal of Verbal Learning \& Verbal Behavior}, vol.~8, no.~2, pp. 240--247, 1969.

\bibitem{smith1974}
E.~E. Smith, E.~J. Shoben, and L.~J. Rips, ``Structure and process in semantic memory: A featural model for semantic decisions,'' \emph{Psychological Review}, vol.~81, no.~3, pp. 214--241, 1974.

\bibitem{Rogers2004}
T.~T. Rogers and J.~L. McClelland, \emph{Semantic Cognition: A Parallel Distributed Processing Approach}.\hskip 1em plus 0.5em minus 0.4em\relax MIT Press, 2004.

\bibitem{yuan2023categorical}
Y.~Yuan, ``A categorical framework of general intelligence,'' 2023, cite arxiv:2303.04571.

\bibitem{riehl2017category}
E.~Riehl, \emph{Category theory in context}, ser. Aurora: {Dover} modern math originals.\hskip 1em plus 0.5em minus 0.4em\relax Dover Publications, 2017.

\bibitem{rissanen1978mdlorigin}
J.~Rissanen, ``Modeling by shortest data description,'' \emph{Automatica}, vol.~14, no.~5, pp. 465--471, 1978.

\bibitem{diederik2024aevb}
D.~P. Kingma and M.~Welling, ``Auto-encoding variational bayes,'' in \emph{Proc.~Int'l Conf.~Learning Representations}, 2014.

\bibitem{blum2003lkm}
A.~Blum and J.~Langford, ``Pac-mdl bounds,'' in \emph{Learning Theory and Kernel Machines}, B.~Sch{\"o}lkopf and M.~K. Warmuth, Eds.\hskip 1em plus 0.5em minus 0.4em\relax Berlin, Heidelberg: Springer Berlin Heidelberg, 2003, pp. 344--357.

\bibitem{sefidgaran2023minimum}
\BIBentryALTinterwordspacing
M.~Sefidgaran, A.~Zaidi, and P.~Krasnowski, ``Minimum description length and generalization guarantees for representation learning,'' in \emph{Thirty-seventh Conference on Neural Information Processing Systems}, 2023. [Online]. Available: \url{https://openreview.net/forum?id=Ncb0MvVqRV}
\BIBentrySTDinterwordspacing

\bibitem{heuse2017fidl}
M.~Heusel, H.~Ramsauer, T.~Unterthiner, B.~Nessler, and S.~Hochreiter, ``Gans trained by a two time-scale update rule converge to a local nash equilibrium,'' in \emph{Proc.~Annual Conf.~Neural Information Processing Systems}, 2017, p. 6629–6640.

\bibitem{hu2018pro}
Y.~Hu, W.~Yang, M.~Li, and J.~Liu, ``Progressive spatial recurrent neural network for intra prediction,'' \emph{{IEEE} Trans. on Multimedia}, vol.~21, no.~12, pp. 3024--3037, 2019.

\bibitem{hochreiter1997lstm}
S.~Hochreiter and J.~Schmidhuber, ``Long short-term memory,'' \emph{Neural Computation}, vol.~9, no.~8, pp. 1735--1780, 1997.

\bibitem{ho2020denoising}
J.~Ho, A.~Jain, and P.~Abbeel, ``Denoising diffusion probabilistic models,'' in \emph{Proc.~Annual Conf.~Neural Information Processing Systems}, 2020.

\bibitem{gao2023rag}
Y.~{Gao}, Y.~{Xiong}, X.~{Gao}, K.~{Jia}, J.~{Pan}, Y.~{Bi}, Y.~{Dai}, J.~{Sun}, M.~{Wang}, and H.~{Wang}, ``{Retrieval-Augmented Generation for Large Language Models: A Survey},'' \emph{arXiv e-prints}, p. arXiv:2312.10997, December 2023.

\bibitem{alex2017alexnet}
A.~Krizhevsky, I.~Sutskever, and G.~E. Hinton, ``Imagenet classification with deep convolutional neural networks,'' \emph{Commun. ACM}, vol.~60, no.~6, p. 84–90, may 2017.

\bibitem{kingma2019vae}
D.~P. Kingma and M.~Welling, 2019.

\bibitem{kobyzev2021normal}
I.~Kobyzev, S.~J. Prince, and M.~A. Brubaker, ``Normalizing flows: An introduction and review of current methods,'' \emph{{IEEE} Trans. on Pattern Analysis and Machine Intelligence}, vol.~43, no.~11, pp. 3964--3979, 2021.

\bibitem{kaplan2019siri}
A.~Kaplan and M.~Haenlein, ``Siri, siri, in my hand: Who’s the fairest in the land? on the interpretations, illustrations, and implications of artificial intelligence,'' \emph{Business Horizons}, vol.~62, no.~1, pp. 15--25, 2019.

\bibitem{campbell2002deepblue}
M.~Campbell, A.~Hoane, and F.~hsiung Hsu, ``Deep blue,'' \emph{Artificial Intelligence}, vol. 134, no.~1, pp. 57--83, 2002.

\bibitem{Kismet}
\BIBentryALTinterwordspacing
Kismet. [Online]. Available: \url{http://www.ai.mit.edu/projects/sociable/baby-bits.html}
\BIBentrySTDinterwordspacing

\bibitem{Har28}
R.~V.~L. Hartley, ``Transmission of information,'' \emph{Bell Syst. Tech. Journal}, vol.~7, pp. 535--563, 1928.

\bibitem{shannon}
C.~E. Shannon, ``A mathematical theory of communication,'' \emph{The Bell System Technical Journal}, vol.~27, pp. 379--423, 1948.

\bibitem{huffman}
D.~A. Huffman, ``A method for the construction of minimum-redundancy codes,'' \emph{Proceedings of the IRE}, vol.~40, no.~9, pp. 1098--1101, 1952.

\bibitem{witten1987arithmetic}
I.~H. Witten, R.~M. Neal, and J.~G. Cleary, ``Arithmetic coding for data compression,'' \emph{Commun. ACM}, vol.~30, no.~6, p. 520–540, jun 1987.

\bibitem{pohlig1980ft}
S.~Pohlig, ``Fourier transform phase coding of images,'' \emph{{IEEE} Trans. on Acoustics, Speech and Signal Processing}, vol.~28, no.~3, pp. 339--341, 1980.

\bibitem{ahmed1974dct}
N.~Ahmed, T.~Natarajan, and K.~Rao, ``Discrete cosine transform,'' \emph{{IEEE} Trans. on Communications}, vol. C-23, no.~1, pp. 90--93, 1974.

\bibitem{wallace1992jpeg}
G.~Wallace, ``The jpeg still picture compression standard,'' \emph{{IEEE} Trans. on Consumer Electronics}, vol.~38, no.~1, pp. xviii--xxxiv, 1992.

\bibitem{kalva2006h264}
H.~Kalva, ``The h.264 video coding standard,'' \emph{IEEE MultiMedia}, vol.~13, no.~4, pp. 86--90, 2006.

\bibitem{fan2004avs}
L.~Fan, S.~Ma, and F.~Wu, ``Overview of {AVS} video standard,'' in \emph{Proc.~IEEE Int'l Conf.~Multimedia and Expo}, vol.~1, 2004, pp. 423--426 Vol.1.

\bibitem{sullivan2012hevc}
G.~J. Sullivan, J.-R. Ohm, W.-J. Han, and T.~Wiegand, ``Overview of the high efficiency video coding ({HEVC}) standard,'' \emph{{IEEE} Trans. on Circuits and Systems for Video Technology}, vol.~22, no.~12, pp. 1649--1668, 2012.

\bibitem{park2016inloop}
W.-S. Park and M.~Kim, ``{CNN}-based in-loop filtering for coding efficiency improvement,'' in \emph{Proc. of IEEE Image, Video, and Multidimensional Signal Processing Workshop (IVMSP)}, 2016, pp. 1--5.

\bibitem{liu2016cupar}
Z.~Liu, X.~Yu, Y.~Gao, S.~Chen, X.~Ji, and D.~Wang, ``{CU} partition mode decision for hevc hardwired intra encoder using convolution neural network,'' \emph{{IEEE} Trans. on Image Processing}, vol.~25, no.~11, pp. 5088--5103, 2016.

\bibitem{li2018line}
J.~Li, B.~Li, J.~Xu, and R.~Xiong, ``Efficient multiple-line-based intra prediction for {HEVC},'' \emph{{IEEE} Trans. on Circuits and Systems for Video Technology}, vol.~28, no.~4, pp. 947--957, 2018.

\bibitem{xia2018group}
S.~Xia, W.~Yang, Y.~Hu, S.~Ma, and J.~Liu, ``A group variational transformation neural network for fractional interpolation of video coding,'' in \emph{Proc.~Data Compression Conference}, 2018, pp. 127--136.

\bibitem{dai2018cuclass}
Y.~Dai, D.~Liu, Z.-J. Zha, and F.~Wu, ``A {CNN}-based in-loop filter with cu classification for {HEVC},'' in \emph{Proc.~IEEE~Visual Communication and Image Processing}, 2018, pp. 1--4.

\bibitem{johannes2016gdn}
J.~Ball{\'{e}}, V.~Laparra, and E.~P. Simoncelli, ``Density modeling of images using a generalized normalization transformation,'' in \emph{Proc.~Int'l Conf.~Learning Representations}, 2016.

\bibitem{balle2017endtoend}
J.~Ball{\'e}, V.~Laparra, and E.~P. Simoncelli, ``End-to-end optimized image compression,'' in \emph{Proc.~Int'l Conf.~Learning Representations}, 2017.

\bibitem{toderici2017rnn}
G.~Toderici, D.~Vincent, N.~Johnston, S.~J. Hwang, D.~Minnen, J.~Shor, and M.~Covell, ``Full resolution image compression with recurrent neural networks,'' in \emph{Proc.~IEEE Int'l Conf.~Computer Vision and Pattern Recognition}, 2017, pp. 5435--5443.

\bibitem{cover2006info}
T.~M. Cover and J.~A. Thomas, \emph{Elements of Information Theory (Wiley Series in Telecommunications and Signal Processing)}.\hskip 1em plus 0.5em minus 0.4em\relax USA: Wiley-Interscience, 2006.

\bibitem{xu2019infotheory}
Y.~Xu, P.~Cao, Y.~Kong, and Y.~Wang, \emph{{LDMI}: a novel information-theoretic loss function for training deep nets robust to label noise}, 2019.

\bibitem{csiszar1975divergence}
I.~Csiszar, ``I-divergence geometry of probability distributions and minimization problems,'' \emph{The Annals of Probability}, vol.~3, no.~1, pp. 146 -- 158, 1975.

\bibitem{lin2013network}
M.~Lin, Q.~Chen, and S.~Yan, ``Network in network,'' 2013.

\bibitem{Ronneberger2015unet}
O.~Ronneberger, P.~Fischer, and T.~Brox, ``U-net: Convolutional networks for biomedical image segmentation,'' in \emph{Medical Image Computing and Computer-Assisted Intervention}, 2015, pp. 234--241.

\bibitem{srivastava2014dropout}
N.~Srivastava, G.~Hinton, A.~Krizhevsky, I.~Sutskever, and R.~Salakhutdinov, ``Dropout: a simple way to prevent neural networks from overfitting,'' \emph{Journal of Machine Learning Research}, vol.~15, no.~1, p. 1929–1958, jan 2014.

\bibitem{he2015spp}
K.~He, X.~Zhang, S.~Ren, and J.~Sun, ``Spatial pyramid pooling in deep convolutional networks for visual recognition,'' \emph{{IEEE} Trans. on Pattern Analysis and Machine Intelligence}, vol.~37, no.~9, pp. 1904--1916, 2015.

\bibitem{he2022mae}
K.~He, X.~Chen, S.~Xie, Y.~Li, P.~Dollár, and R.~Girshick, ``Masked autoencoders are scalable vision learners,'' in \emph{Proc.~IEEE Int'l Conf.~Computer Vision and Pattern Recognition}, 2022, pp. 15\,979--15\,988.

\bibitem{michael2018on}
A.~M. Saxe, Y.~Bansal, J.~Dapello, M.~Advani, A.~Kolchinsky, B.~D. Tracey, and D.~D. Cox, ``On the information bottleneck theory of deep learning,'' in \emph{Proc.~Int'l Conf.~Learning Representations}, 2018.

\bibitem{naftali2015ibp}
N.~Tishby and N.~Zaslavsky, ``Deep learning and the information bottleneck principle,'' \emph{arXiv}, vol. abs/1503.02406, 2015.

\bibitem{icml2023kzxinfodl}
K.~Kawaguchi, Z.~Deng, X.~Ji, and J.~Huang, ``How does information bottleneck help deep learning?'' in \emph{Proc.~Int'l Conf.~Machine Learning}, 2023.

\bibitem{yu2020learning}
Y.~Yu, K.~H.~R. Chan, C.~You, C.~Song, and Y.~Ma, ``Learning diverse and discriminative representations via the principle of maximal coding rate reduction,'' \emph{Proc.~Annual Conf.~Neural Information Processing Systems}, vol.~33, 2020.

\bibitem{lai2019lpn}
W.-S. Lai, J.-B. Huang, N.~Ahuja, and M.-H. Yang, ``Fast and accurate image super-resolution with deep laplacian pyramid networks,'' \emph{{IEEE} Trans. on Pattern Analysis and Machine Intelligence}, vol.~41, no.~11, pp. 2599--2613, 2019.

\bibitem{wang2004ssim}
Z.~Wang, A.~Bovik, H.~Sheikh, and E.~Simoncelli, ``Image quality assessment: from error visibility to structural similarity,'' \emph{{IEEE} Trans. on Image Processing}, vol.~13, no.~4, pp. 600--612, 2004.

\bibitem{zhang2018lpips}
R.~Zhang, P.~Isola, A.~A. Efros, E.~Shechtman, and O.~Wang, ``The unreasonable effectiveness of deep features as a perceptual metric,'' in \emph{Proc.~IEEE Int'l Conf.~Computer Vision and Pattern Recognition}, 2018, pp. 586--595.

\bibitem{ding2022dist}
K.~Ding, K.~Ma, S.~Wang, and E.~P. Simoncelli, ``Image quality assessment: Unifying structure and texture similarity,'' \emph{{IEEE} Trans. on Pattern Analysis and Machine Intelligence}, vol.~44, no.~5, pp. 2567--2581, 2022.

\bibitem{balcan2016softmax}
W.~Liu, Y.~Wen, Z.~Yu, and M.~Yang, ``Large-margin softmax loss for convolutional neural networks,'' in \emph{Proc.~Int'l Conf.~Machine Learning}, ser. Proceedings of Machine Learning Research, vol.~48, Jun 2016, pp. 507--516.

\bibitem{mao2023cross_entropy}
A.~Mao, M.~Mohri, and Y.~Zhong, ``Cross-entropy loss functions: theoretical analysis and applications,'' in \emph{Proc.~Int'l Conf.~Machine Learning}, 2023.

\bibitem{luo2021shl}
J.~Luo, H.~Qiao, and B.~Zhang, ``Learning with smooth hinge losses,'' \emph{Neurocomputing}, vol. 463, pp. 379--387, 2021.

\bibitem{zhou2019iou}
D.~Zhou, J.~Fang, X.~Song, C.~Guan, J.~Yin, Y.~Dai, and R.~Yang, ``Iou loss for 2d/3d object detection,'' in \emph{Proc.~Int'l Conf. on 3D Vision (3DV)}, 2019, pp. 85--94.

\bibitem{lin2017focal}
T.-Y. Lin, P.~Goyal, R.~Girshick, K.~He, and P.~Dollár, ``Focal loss for dense object detection,'' in \emph{Proc.~IEEE Int'l Conf.~Computer Vision}, 2017, pp. 2999--3007.

\bibitem{text8_dataset}
\BIBentryALTinterwordspacing
Text8 dataset. [Online]. Available: \url{https://mattmahoney.net/dc/textdata.html}
\BIBentrySTDinterwordspacing

\bibitem{gutenberg}
Denton, \emph{Legends of Texas}.\hskip 1em plus 0.5em minus 0.4em\relax University of North Texas Press.

\bibitem{valmeekam2023llmzip}
C.~S.~K. Valmeekam, K.~Narayanan, D.~Kalathil, J.-F. Chamberland, and S.~Shakkottai, ``Llmzip: Lossless text compression using large language models,'' 2023.

\bibitem{gemmeke2017audio}
J.~F. Gemmeke, D.~P.~W. Ellis, D.~Freedman, A.~Jansen, W.~Lawrence, R.~C. Moore, M.~Plakal, and M.~Ritter, ``Audio set: An ontology and human-labeled dataset for audio events,'' in \emph{Proc.~IEEE Int'l Conf.~Acoustics, Speech, and Signal Processing}, 2017, pp. 776--780.

\bibitem{mysore2015audio}
G.~J. Mysore, ``Can we automatically transform speech recorded on common consumer devices in real-world environments into professional production quality speech?—a dataset, insights, and challenges,'' \emph{{IEEE} Signal Processing Letters}, vol.~22, no.~8, pp. 1006--1010, 2015.

\bibitem{musdb18}
\BIBentryALTinterwordspacing
Musdb18. [Online]. Available: \url{https://source-separation.github.io/tutorial/data/musdb18.html}
\BIBentrySTDinterwordspacing

\bibitem{kumar2023highfidelity}
R.~Kumar, P.~Seetharaman, A.~Luebs, I.~Kumar, and K.~Kumar, ``High-fidelity audio compression with improved rvqgan,'' 2023.

\bibitem{cdvs_dataset}
\BIBentryALTinterwordspacing
Compact descriptors for visual search {(CDVS)}. [Online]. Available: \url{https://exhibits.stanford.edu/data/catalog/qy869qz5226}
\BIBentrySTDinterwordspacing

\bibitem{duan2015cdvs}
L.-Y. Duan, T.~Huang, and W.~Gao, ``Overview of the mpeg cdvs standard,'' in \emph{Proc.~Data Compression Conference}, 2015, pp. 323--332.

\bibitem{liu2016cvpr}
H.~Liu, Y.~Tian, Y.~Wang, L.~Pang, and T.~Huang, ``Deep relative distance learning: Tell the difference between similar vehicles,'' in \emph{Proc.~IEEE Int'l Conf.~Computer Vision and Pattern Recognition}, 2016, pp. 2167--2175.

\bibitem{bai2022veriwild2}
Y.~Bai, J.~Liu, Y.~Lou, C.~Wang, and L.-Y. Duan, ``Disentangled feature learning network and a comprehensive benchmark for vehicle re-identification,'' \emph{{IEEE} Trans. on Pattern Analysis and Machine Intelligence}, vol.~44, no.~10, pp. 6854--6871, 2022.

\bibitem{zheng2015market1501}
L.~Zheng, L.~Shen, L.~Tian, S.~Wang, J.~Wang, and Q.~Tian, ``Scalable person re-identification: A benchmark,'' in \emph{Proc.~IEEE Int'l Conf.~Computer Vision}, 2015, pp. 1116--1124.

\bibitem{duan2018tip}
L.-y. Duan, Y.~Wu, Y.~Huang, Z.~Wang, J.~Yuan, and W.~Gao, ``Minimizing reconstruction bias hashing via joint projection learning and quantization,'' \emph{{IEEE} Trans. on Image Processing}, vol.~27, no.~6, pp. 3127--3141, 2018.

\bibitem{deng2009imagenet}
J.~Deng, W.~Dong, R.~Socher, L.-J. Li, K.~Li, and L.~Fei-Fei, ``Imagenet: A large-scale hierarchical image database,'' in \emph{Proc.~IEEE Int'l Conf.~Computer Vision and Pattern Recognition}, 2009, pp. 248--255.

\bibitem{chen2020intermediate}
Z.~Chen, K.~Fan, S.~Wang, L.~Duan, W.~Lin, and A.~C. Kot, ``Toward intelligent sensing: Intermediate deep feature compression,'' \emph{{IEEE} Trans. on Image Processing}, vol.~29, pp. 2230--2243, 2020.

\bibitem{lin2014coco}
T.-Y. Lin, M.~Maire, S.~Belongie, J.~Hays, P.~Perona, D.~Ramanan, P.~Doll{\'a}r, and C.~L. Zitnick, ``Microsoft coco: Common objects in context,'' in \emph{Proc.~IEEE European Conf.~Computer Vision}, 2014, pp. 740--755.

\bibitem{choi2022scalable}
H.~Choi and I.~V. Bajić, ``Scalable image coding for humans and machines,'' \emph{{IEEE} Trans. on Image Processing}, vol.~31, pp. 2739--2754, 2022.

\bibitem{cordts2016cityscapes}
M.~Cordts, M.~Omran, S.~Ramos, T.~Rehfeld, M.~Enzweiler, R.~Benenson, U.~Franke, S.~Roth, and B.~Schiele, ``The cityscapes dataset for semantic urban scene understanding,'' in \emph{Proc.~IEEE Int'l Conf.~Computer Vision and Pattern Recognition}, 2016, pp. 3213--3223.

\bibitem{yang2020pcc}
R.~Yang, N.~Yan, L.~Li, D.~Liu, and F.~Wu, ``Chain code-based occupancy map coding for video-based point cloud compression,'' in \emph{Proc.~IEEE~Visual Communication and Image Processing}, 2020, pp. 479--482.

\bibitem{zamir2018taskonomy}
A.~R. Zamir, A.~Sax, W.~Shen, L.~Guibas, J.~Malik, and S.~Savarese, ``Taskonomy: Disentangling task transfer learning,'' in \emph{Proc.~IEEE Int'l Conf.~Computer Vision and Pattern Recognition}, 2018, pp. 3712--3722.

\bibitem{lin2020HumanIE}
W.~Lin, H.~Liu, S.~Liu, Y.~Li, G.-J. Qi, R.~Qian, T.~Wang, N.~Sebe, N.~Xu, H.~Xiong, and M.~Shah, ``Human in events: A large-scale benchmark for human-centric video analysis in complex events,'' \emph{Int'l Journal of Computer Vision}, vol. 131, pp. 2994--3018, 2023.

\bibitem{lin2020pointcompress}
W.~Lin, X.~He, W.~Dai, J.~See, T.~Shinde, H.~Xiong, and L.~Duan, ``Key-point sequence lossless compression for intelligent video analysis,'' \emph{IEEE MultiMedia}, vol.~27, no.~3, pp. 12--22, 2020.

\bibitem{clic}
\BIBentryALTinterwordspacing
Workshop and challenge on learned image compression. [Online]. Available: \url{http://www.compression.cc}
\BIBentrySTDinterwordspacing

\bibitem{wang2022neuraltrans}
D.~Wang, W.~Yang, Y.~Hu, and J.~Liu, ``Neural data-dependent transform for learned image compression,'' in \emph{Proc.~IEEE Int'l Conf.~Computer Vision and Pattern Recognition}, 2022, pp. 17\,358--17\,367.

\bibitem{mentzer2020high}
F.~Mentzer, G.~Toderici, M.~Tschannen, and E.~Agustsson, ``High-fidelity generative image compression,'' in \emph{Proc.~Annual Conf.~Neural Information Processing Systems}, 2020.

\bibitem{orel2020lifespan}
R.~Or-El, S.~Sengupta, O.~Fried, E.~Shechtman, and I.~Kemelmacher-Shlizerman, ``Lifespan age transformation synthesis,'' in \emph{Proc.~IEEE European Conf.~Computer Vision}, 2020.

\bibitem{yang2023manifold}
W.~Yang, H.~Huang, J.~Liu, and A.~C. Kot, ``Facial image compression via neural image manifold compression,'' \emph{{IEEE} Trans. on Circuits and Systems for Video Technology}, pp. 1--1, 2023.

\bibitem{behley2019SemanticKITTI}
J.~Behley, M.~Garbade, A.~Milioto, J.~Quenzel, S.~Behnke, C.~Stachniss, and J.~Gall, ``Semantickitti: A dataset for semantic scene understanding of lidar sequences,'' in \emph{Proc.~IEEE Int'l Conf.~Computer Vision}, 2019, pp. 9296--9306.

\bibitem{he2022density_pcc}
Y.~He, X.~Ren, D.~Tang, Y.~Zhang, X.~Xue, and Y.~Fu, ``Density-preserving deep point cloud compression,'' in \emph{Proc.~IEEE Int'l Conf.~Computer Vision and Pattern Recognition}, 2022, pp. 2323--2332.

\bibitem{AVIRIS}
\BIBentryALTinterwordspacing
Airborne visible/infrared imaging spectrometer dataset. [Online]. Available: \url{https://aviris.jpl.nasa.gov/}
\BIBentrySTDinterwordspacing

\bibitem{Meyer2020Multispectral}
A.~Meyer, N.~Genser, and A.~Kaup, ``Multispectral image compression based on hevc using pel-recursive inter-band prediction,'' in \emph{Proc. IEEE Int'l Workshop on Multimedia Signal Processing (MMSP)}, 2020, pp. 1--6.

\bibitem{Bao2020InStereo2KAL}
W.~Bao, W.~Wang, Y.~Xu, Y.~Guo, S.~Hong, and X.~Zhang, ``Instereo2k: a large real dataset for stereo matching in indoor scenes,'' \emph{Science China Information Sciences}, vol.~63, 2020.

\bibitem{Wodlinger2022SASIC}
M.~Wödlinger, J.~Kotera, J.~Xu, and R.~Sablatnig, ``Sasic: Stereo image compression with latent shifts and stereo attention,'' in \emph{Proc.~IEEE Int'l Conf.~Computer Vision and Pattern Recognition}, 2022, pp. 651--660.

\bibitem{nih}
\BIBentryALTinterwordspacing
Nih dataset. [Online]. Available: \url{https://www.nih.gov/news-events/news-releases/nih-clinical-center-releases-dataset-32000-ct-images}
\BIBentrySTDinterwordspacing

\bibitem{guo2023xray}
Z.~Guo, S.~Zhao, D.~Han, and C.~Yang, ``X-ray image compression using variational auto-encoder,'' in \emph{Proc. of the Int'l Conf. on Electronic Information Technology and Computer Engineering}, 2023, p. 1277–1281.

\bibitem{wang2020rgbdavis}
Z.~W. Wang, P.~Duan, O.~Cossairt, A.~Katsaggelos, T.~Huang, and B.~Shi, ``Joint filtering of intensity images and neuromorphic events for high-resolution noise-robust imaging,'' in \emph{Proc.~IEEE Int'l Conf.~Computer Vision and Pattern Recognition}, 2020, pp. 1606--1616.

\bibitem{banerjee2021event}
S.~Banerjee, Z.~W. Wang, H.~H. Chopp, O.~Cossairt, and A.~K. Katsaggelos, ``Lossy event compression based on image-derived quad trees and poisson disk sampling,'' in \emph{Proc.~IEEE Int'l Conf.~Image Processing}, 2021, pp. 2154--2158.

\bibitem{nie2023kmean}
F.~Nie, Z.~Li, R.~Wang, and X.~Li, ``An effective and efficient algorithm for k-means clustering with new formulation,'' vol.~35, no.~4, pp. 3433--3443, 2023.

\bibitem{jolliffe2016pca}
I.~T. Jolliffe and J.~Cadima, ``Principal component analysis: a review and recent developments,'' \emph{Philosophical Transactions of the Royal Society A: Mathematical, Physical and Engineering Sciences}, vol. 374, no. 2065, p. 20150202, 2016.

\bibitem{goodfellow2014gan}
I.~J. Goodfellow, J.~Pouget-Abadie, M.~Mirza, B.~Xu, D.~Warde-Farley, S.~Ozair, A.~Courville, and Y.~Bengio, ``Generative adversarial nets,'' in \emph{Proc.~Annual Conf.~Neural Information Processing Systems}, 2014, p. 2672–2680.

\bibitem{martin2017icml}
M.~Arjovsky, S.~Chintala, and L.~Bottou, ``{W}asserstein generative adversarial networks,'' in \emph{Proc.~Int'l Conf.~Machine Learning}, vol.~70, 06--11 Aug 2017, pp. 214--223.

\bibitem{chen2020simclr}
T.~Chen, S.~Kornblith, M.~Norouzi, and G.~Hinton, ``A simple framework for contrastive learning of visual representations,'' in \emph{Proc.~Int'l Conf.~Machine Learning}, 2020.

\bibitem{nowozin2016fgan}
S.~Nowozin, B.~Cseke, and R.~Tomioka, ``f-gan: training generative neural samplers using variational divergence minimization,'' in \emph{Proc.~Annual Conf.~Neural Information Processing Systems}, 2016, p. 271–279.

\bibitem{zhang2022fmutual}
\BIBentryALTinterwordspacing
G.~Zhang, Y.~Lu, S.~Sun, H.~Guo, and Y.~Yu, ``f-mutual information contrastive learning,'' 2022. [Online]. Available: \url{https://openreview.net/forum?id=3kTt_W1_tgw}
\BIBentrySTDinterwordspacing

\bibitem{rosasco2004same}
L.~Rosasco, E.~D. Vito, A.~Caponnetto, M.~Piana, and A.~Verri, ``Are loss functions all the same?'' \emph{Neural Computation}, vol.~16, no.~5, pp. 1063--1076, 2004.

\bibitem{ferrucci2013watson}
D.~Ferrucci, A.~Levas, S.~Bagchi, D.~Gondek, and E.~T. Mueller, ``Watson: Beyond jeopardy!'' \emph{Artificial Intelligence}, vol. 199-200, pp. 93--105, 2013.

\bibitem{wang2018ssunderstanding}
P.~Wang, P.~Chen, Y.~Yuan, D.~Liu, Z.~Huang, X.~Hou, and G.~Cottrell, ``Understanding convolution for semantic segmentation,'' in \emph{Proc.~IEEE Winter Conference on Applications of Computer Vision (WACV)}, 2018, pp. 1451--1460.

\bibitem{laga2022stereodepth}
H.~Laga, L.~V. Jospin, F.~Boussaid, and M.~Bennamoun, ``A survey on deep learning techniques for stereo-based depth estimation,'' \emph{{IEEE} Trans. on Pattern Analysis and Machine Intelligence}, vol.~44, no.~4, pp. 1738--1764, 2022.

\bibitem{poggi2022bsde}
M.~Poggi, F.~Tosi, K.~Batsos, P.~Mordohai, and S.~Mattoccia, ``On the synergies between machine learning and binocular stereo for depth estimation from images: A survey,'' \emph{{IEEE} Trans. on Pattern Analysis and Machine Intelligence}, vol.~44, no.~9, pp. 5314--5334, 2022.

\bibitem{schuster1997brnn}
M.~Schuster and K.~Paliwal, ``Bidirectional recurrent neural networks,'' \emph{{IEEE} Trans. on Signal Processing}, vol.~45, no.~11, pp. 2673--2681, 1997.

\bibitem{liu2021swintrans}
Z.~Liu, Y.~Lin, Y.~Cao, H.~Hu, Y.~Wei, Z.~Zhang, S.~Lin, and B.~Guo, ``Swin transformer: Hierarchical vision transformer using shifted windows,'' in \emph{Proc.~IEEE Int'l Conf.~Computer Vision}, 2021, pp. 9992--10\,002.

\bibitem{han2023vit}
K.~Han, Y.~Wang, H.~Chen, X.~Chen, J.~Guo, Z.~Liu, Y.~Tang, A.~Xiao, C.~Xu, Y.~Xu, Z.~Yang, Y.~Zhang, and D.~Tao, ``A survey on vision transformer,'' \emph{{IEEE} Trans. on Pattern Analysis and Machine Intelligence}, vol.~45, no.~1, pp. 87--110, 2023.

\bibitem{devlin2019bert}
J.~Devlin, M.-W. Chang, K.~Lee, and K.~Toutanova, ``{BERT}: Pre-training of deep bidirectional transformers for language understanding,'' in \emph{Proc. of the Conference of the North {A}merican Chapter of the Association for Computational Linguistics: Human Language Technologies, Volume 1 (Long and Short Papers)}, June 2019, pp. 4171--4186.

\bibitem{bao2022vlmo}
H.~Bao, W.~Wang, L.~Dong, Q.~Liu, O.~K. Mohammed, K.~Aggarwal, S.~Som, S.~Piao, and F.~Wei, ``Vlmo: Unified vision-language pre-training with mixture-of-modality-experts,'' in \emph{Proc.~Annual Conf.~Neural Information Processing Systems}, vol.~35, 2022, pp. 32\,897--32\,912.

\bibitem{dhariwal2021diffusion}
P.~Dhariwal and A.~Nichol, ``Diffusion models beat gans on image synthesis,'' in \emph{Proc.~Annual Conf.~Neural Information Processing Systems}, vol.~34, 2021, pp. 8780--8794.

\bibitem{yang2023diffusuion}
L.~Yang, Z.~Zhang, Y.~Song, S.~Hong, R.~Xu, Y.~Zhao, W.~Zhang, B.~Cui, and M.-H. Yang, ``Diffusion models: A comprehensive survey of methods and applications,'' \emph{ACM Comput. Surv.}, vol.~56, no.~4, nov 2023.

\bibitem{le2022deeprl}
N.~Le, V.~S. Rathour, K.~Yamazaki, K.~Luu, and M.~Savvides, ``Deep reinforcement learning in computer vision: a comprehensive survey,'' \emph{Artif. Intell. Rev.}, vol.~55, no.~4, p. 2733–2819, apr 2022.

\bibitem{sutton2018rl}
R.~S. Sutton and A.~G. Barto, \emph{Reinforcement Learning: An Introduction}.\hskip 1em plus 0.5em minus 0.4em\relax Cambridge, MA, USA: A Bradford Book, 2018.

\bibitem{creswell2018ganoverview}
A.~Creswell, T.~White, V.~Dumoulin, K.~Arulkumaran, B.~Sengupta, and A.~A. Bharath, ``Generative adversarial networks: An overview,'' \emph{{IEEE} Signal Processing Magazine}, vol.~35, no.~1, pp. 53--65, 2018.

\bibitem{goodfellow2020}
I.~Goodfellow, J.~Pouget-Abadie, M.~Mirza, B.~Xu, D.~Warde-Farley, S.~Ozair, A.~Courville, and Y.~Bengio, ``Generative adversarial networks,'' \emph{Commun. ACM}, vol.~63, no.~11, p. 139–144, oct 2020.

\bibitem{rombach2022sd}
R.~Rombach, A.~Blattmann, D.~Lorenz, P.~Esser, and B.~Ommer, ``High-resolution image synthesis with latent diffusion models,'' in \emph{Proc.~IEEE Int'l Conf.~Computer Vision and Pattern Recognition}, 2022, pp. 10\,674--10\,685.

\bibitem{dalle}
\BIBentryALTinterwordspacing
Dall-e. [Online]. Available: \url{https://openai.com/dall-e-3}
\BIBentrySTDinterwordspacing

\bibitem{chatgpt}
\BIBentryALTinterwordspacing
Chatgpt. [Online]. Available: \url{https://chat.openai.com/}
\BIBentrySTDinterwordspacing

\bibitem{midjourney}
\BIBentryALTinterwordspacing
midjourney. [Online]. Available: \url{https://www.midjourney.com/home}
\BIBentrySTDinterwordspacing

\bibitem{sora}
\BIBentryALTinterwordspacing
Sora. [Online]. Available: \url{https://openai.com/research/video-generation-models-as-world-simulators}
\BIBentrySTDinterwordspacing

\bibitem{kawar2023imagic}
B.~Kawar, S.~Zada, O.~Lang, O.~Tov, H.~Chang, T.~Dekel, I.~Mosseri, and M.~Irani, ``Imagic: Text-based real image editing with diffusion models,'' in \emph{Proc.~IEEE Int'l Conf.~Computer Vision and Pattern Recognition}, 2023, pp. 6007--6017.

\bibitem{zhang2022mtl}
Y.~Zhang and Q.~Yang, ``A survey on multi-task learning,'' vol.~34, no.~12, pp. 5586--5609, 2022.

\bibitem{gabriela2017dava}
G.~Csurka, ``Domain adaptation for visual applications: {A} comprehensive survey,'' \emph{arXiv}, vol. abs/1702.05374, 2017.

\bibitem{wang2018sft}
X.~Wang, K.~Yu, C.~Dong, and C.~Change~Loy, ``Recovering realistic texture in image super-resolution by deep spatial feature transform,'' in \emph{Proc.~IEEE Int'l Conf.~Computer Vision and Pattern Recognition}, 2018, pp. 606--615.

\bibitem{zhong2023gsr}
Z.~Zhong, X.~Liu, J.~Jiang, D.~Zhao, and X.~Ji, ``Guided depth map super-resolution: A survey,'' \emph{ACM Comput. Surv.}, vol.~55, no. 14s, jul 2023.

\bibitem{li2022allinone}
B.~Li, X.~Liu, P.~Hu, Z.~Wu, J.~Lv, and X.~Peng, ``All-in-one image restoration for unknown corruption,'' in \emph{Proc.~IEEE Int'l Conf.~Computer Vision and Pattern Recognition}, 2022, pp. 17\,431--17\,441.

\bibitem{vandenhende2022mtldense}
S.~Vandenhende, S.~Georgoulis, W.~Van~Gansbeke, M.~Proesmans, D.~Dai, and L.~Van~Gool, ``Multi-task learning for dense prediction tasks: A survey,'' \emph{{IEEE} Trans. on Pattern Analysis and Machine Intelligence}, vol.~44, no.~7, pp. 3614--3633, 2022.

\bibitem{szegedy2015googlenet}
C.~Szegedy, W.~Liu, Y.~Jia, P.~Sermanet, S.~Reed, D.~Anguelov, D.~Erhan, V.~Vanhoucke, and A.~Rabinovich, ``Going deeper with convolutions,'' in \emph{Proc.~IEEE Int'l Conf.~Computer Vision and Pattern Recognition}, 2015, pp. 1--9.

\bibitem{xiong2020ln}
R.~Xiong, Y.~Yang, D.~He, K.~Zheng, S.~Zheng, C.~Xing, H.~Zhang, Y.~Lan, L.~Wang, and T.-Y. Liu, ``On layer normalization in the transformer architecture,'' 2020.

\bibitem{Le2021icassp}
N.~Le, H.~Zhang, F.~Cricri, R.~Ghaznavi-Youvalari, and E.~Rahtu, ``Image coding for machines: an end-to-end learned approach,'' in \emph{Proc.~IEEE Int'l Conf.~Acoustics, Speech, and Signal Processing}, 2021, pp. 1590--1594.

\bibitem{Choi2020eccv}
J.~Choi and B.~Han, ``Task-aware quantization network for jpeg image compression,'' in \emph{Proc.~IEEE European Conf.~Computer Vision}, 2020, pp. 309--324.

\bibitem{Alvar2020icassp}
S.~R. Alvar and I.~V. Bajić, ``Bit allocation for multi-task collaborative intelligence,'' in \emph{Proc.~IEEE Int'l Conf.~Acoustics, Speech, and Signal Processing}, 2020, pp. 4342--4346.

\bibitem{Shah2020icassp}
M.~A. Shah and B.~Raj, ``Deriving compact feature representations via annealed contraction,'' in \emph{Proc.~IEEE Int'l Conf.~Acoustics, Speech, and Signal Processing}, 2020, pp. 2068--2072.

\bibitem{wang2021tmm}
S.~Wang, S.~Wang, W.~Yang, X.~Zhang, S.~Wang, S.~Ma, and W.~Gao, ``Towards analysis-friendly face representation with scalable feature and texture compression,'' \emph{{IEEE} Trans. on Multimedia}, pp. 1--1, 2021.

\bibitem{liu2021scalable}
K.~{Liu}, D.~{Liu}, L.~{Li}, N.~{Yan}, and H.~{Li}, ``{Semantics-to-Signal Scalable Image Compression with Learned Revertible Representations},'' \emph{Int'l Journal of Computer Vision}, p. 9/2021, 2021.

\bibitem{yang2021tmm}
S.~Yang, Y.~Hu, W.~Yang, L.-Y. Duan, and J.~Liu, ``Towards coding for human and machine vision: Scalable face image coding,'' \emph{{IEEE} Trans. on Multimedia}, vol.~23, pp. 2957--2971, 2021.

\bibitem{minnen2018ar}
D.~Minnen, J.~Ball\'{e}, and G.~Toderici, ``Joint autoregressive and hierarchical priors for learned image compression,'' in \emph{Proc.~Annual Conf.~Neural Information Processing Systems}, 2018, p. 10794–10803.

\bibitem{Cheng2020}
Z.~Cheng, H.~Sun, M.~Takeuchi, and J.~Katto, ``Learned image compression with discretized gaussian mixture likelihoods and attention modules,'' in \emph{IEEE/CVF Conference on Computer Vision and Pattern Recognition (CVPR)}, 2020, pp. 7936--7945.

\bibitem{hu2020coarse}
Y.~Hu, W.~Yang, and J.~Liu, ``Coarse-to-fine hyper-prior modeling for learned image compression,'' in \emph{Proc.~AAAI Conf. on Artificial Intelligence}, 2020.

\bibitem{lee2019context}
J.~Lee, S.~Cho, and S.-K. Beack, ``Context-adaptive entropy model for end-to-end optimized image compression,'' in \emph{Proc.~Int'l Conf.~Learning Representations}, May 2019.

\bibitem{cao2018vggface2}
Q.~Cao, L.~Shen, W.~Xie, O.~M. Parkhi, and A.~Zisserman, ``Vggface2: A dataset for recognising faces across pose and age,'' in \emph{Proc. of IEEE Int'l Conf. on Automatic Face and Gesture Recognition}, may 2018, pp. 67--74.

\bibitem{jpegai}
\BIBentryALTinterwordspacing
Jpeg ai dataset. [Online]. Available: \url{https://jpeg.org/jpegai/dataset.html}
\BIBentrySTDinterwordspacing

\bibitem{kodak}
\BIBentryALTinterwordspacing
Kodak lossless true color image suite ({PhotoCD} {PCD0992}). [Online]. Available: \url{http://r0k.us/graphics/kodak/}
\BIBentrySTDinterwordspacing

\bibitem{hu2022lora}
E.~J. Hu, Y.~Shen, P.~Wallis, Z.~Allen-Zhu, Y.~Li, S.~Wang, L.~Wang, and W.~Chen, ``Lo{RA}: Low-rank adaptation of large language models,'' in \emph{Proc.~Int'l Conf.~Learning Representations}, 2022.

\bibitem{ding2023peft}
N.~Ding, Y.~Qin, G.~Yang, F.~Wei, Z.~Yang, Y.~Su, S.~Hu, Y.~Chen, C.-M. Chan, W.~Chen, J.~Yi, W.~Zhao, X.~Wang, Z.~Liu, H.-T. Zheng, J.~Chen, Y.~Liu, J.~Tang, J.~Li, and M.~Sun, ``Parameter-efficient fine-tuning of large-scale pre-trained language models,'' \emph{Nature Machine Intelligence}, vol.~5, no.~3, pp. 220--235, March 2023.

\bibitem{Khattak_2023_ICCV}
M.~U. Khattak, S.~T. Wasim, M.~Naseer, S.~Khan, M.-H. Yang, and F.~S. Khan, ``Self-regulating prompts: Foundational model adaptation without forgetting,'' in \emph{Proc.~IEEE Int'l Conf.~Computer Vision}, October 2023, pp. 15\,190--15\,200.

\bibitem{li2023common}
A.~Li, L.~Zhuang, S.~Fan, and S.~Wang, ``Learning common and specific visual prompts for domain generalization,'' in \emph{Proc.~IEEE Asia Conf.~Computer Vision}, 2023, p. 578–593.

\bibitem{chamain2021end}
L.~D. Chamain, F.~Racap{\'e}, J.~B{\'e}gaint, A.~Pushparaja, and S.~Feltman, ``End-to-end optimized image compression for machines, a study,'' in \emph{Proc. of IEEE Data Compression Conference}, 2021.

\bibitem{le2021ICM}
N.~Le, H.~Zhang, F.~Cricri, R.~Ghaznavi-Youvalari, H.~R. Tavakoli, and E.~Rahtu, ``Learned image coding for machines: A content-adaptive approach,'' in \emph{Proc.~IEEE Int'l Conf.~Multimedia and Expo}, 2021, pp. 1--6.

\bibitem{huang2021RDICM}
Z.~Huang, C.~Jia, S.~Wang, and S.~Ma, ``Visual analysis motivated rate-distortion model for image coding,'' in \emph{Proc.~IEEE Int'l Conf.~Multimedia and Expo}, 2021, pp. 1--6.

\bibitem{chen2019lossy}
Z.~Chen, K.~Fan, S.~Wang, L.-Y. Duan, W.~Lin, and A.~Kot, ``Lossy intermediate deep learning feature compression and evaluation,'' in \emph{{ACM} Trans.~Multimedia}, 2019, p. 2414–2422.

\bibitem{alvar2021bitallo}
S.~R. Alvar and I.~V. Bajić, ``Pareto-optimal bit allocation for collaborative intelligence,'' \emph{{IEEE} Trans. on Image Processing}, vol.~30, pp. 3348--3361, 2021.

\bibitem{zhang2021msfc}
Z.~Zhang, M.~Wang, M.~Ma, J.~Li, and X.~Fan, ``{MSFC}: Deep feature compression in multi-task network,'' in \emph{Proc.~IEEE Int'l Conf.~Multimedia and Expo}, 2021, pp. 1--6.

\bibitem{Prabhakar2021dcc}
R.~Prabhakar, S.~Chandak, C.~Chiu, R.~Liang, H.~Nguyen, K.~Tatwawadi, and T.~Weissman, ``Reducing latency and bandwidth for video streaming using keypoint extraction and digital puppetry,'' in \emph{Proc.~Data Compression Conference}, 2021, pp. 360--360.

\bibitem{huh2024platonic}
M.~{Huh}, B.~{Cheung}, T.~{Wang}, and P.~{Isola}, ``{The Platonic Representation Hypothesis},'' \emph{arXiv e-prints}, p. arXiv:2405.07987, May 2024.

\bibitem{zhou2022allloss}
J.~{Zhou}, C.~{You}, X.~{Li}, K.~{Liu}, S.~{Liu}, Q.~{Qu}, and Z.~{Zhu}, ``{Are All Losses Created Equal: A Neural Collapse Perspective},'' \emph{arXiv e-prints}, p. arXiv:2210.02192, October 2022.

\bibitem{blau2018pdt}
Y.~Blau and T.~Michaeli, ``The perception-distortion tradeoff,'' in \emph{Proc.~IEEE Int'l Conf.~Computer Vision and Pattern Recognition}, 2018, pp. 6228--6237.

\bibitem{cohen2024the}
\BIBentryALTinterwordspacing
R.~Cohen, E.~Rivlin, and D.~Freedman, ``The uncertainty-perception tradeoff,'' 2024. [Online]. Available: \url{https://openreview.net/forum?id=sJAlw561AH}
\BIBentrySTDinterwordspacing

\bibitem{brown2020gpt}
T.~Brown, B.~Mann, N.~Ryder, M.~Subbiah, J.~D. Kaplan, P.~Dhariwal, A.~Neelakantan, P.~Shyam, G.~Sastry, A.~Askell, S.~Agarwal, A.~Herbert-Voss, G.~Krueger, T.~Henighan, R.~Child, A.~Ramesh, D.~Ziegler, J.~Wu, C.~Winter, C.~Hesse, M.~Chen, E.~Sigler, M.~Litwin, S.~Gray, B.~Chess, J.~Clark, C.~Berner, S.~McCandlish, A.~Radford, I.~Sutskever, and D.~Amodei, ``Language models are few-shot learners,'' in \emph{Advances in Neural Information Processing Systems}, vol.~33, 2020, pp. 1877--1901.

\bibitem{tian2024var}
K.~{Tian}, Y.~{Jiang}, Z.~{Yuan}, B.~{Peng}, and L.~{Wang}, ``{Visual Autoregressive Modeling: Scalable Image Generation via Next-Scale Prediction},'' \emph{arXiv e-prints}, p. arXiv:2404.02905, April 2024.

\end{thebibliography}



\end{document}